\documentclass{article}

\usepackage{microtype}
\usepackage{graphicx}
\usepackage{subcaption}
\usepackage{booktabs} 
\usepackage{duckuments}

\usepackage{framed}
\usepackage{caption}
\usepackage{svg}
\usepackage{multirow}
\usepackage{multicol}
\usepackage{array}
\usepackage{longtable}
\usepackage{pythonhighlight}
\usepackage{color, colortbl}
\usepackage{xcolor}
\definecolor{LightCyan}{rgb}{0.8, 0.9, 1}
\usepackage{mathtools}
\usepackage{wrapfig}
\usepackage{float} 
\usepackage{tcolorbox}
\usepackage[utf8]{inputenc} 
\usepackage[T1]{fontenc}    
\usepackage{booktabs}       
\usepackage{amsfonts}       
\usepackage{nicefrac}       
\usepackage{microtype}      
\usepackage{xcolor}         
\usepackage{graphicx}
\usepackage{xspace}
\usepackage{amsmath}
\usepackage{enumitem}
\usepackage{adjustbox}
\usepackage{bbm}
\usepackage{amssymb}
\usepackage{pifont}
\usepackage{graphicx}
\usepackage{subcaption}
\usepackage[frozencache,cachedir=.]{minted}
\usepackage{enumitem}
\usepackage{booktabs}
\usepackage{array}
\usepackage{makecell}
\usepackage[table]{xcolor}
\usepackage{caption}
\usepackage{algpseudocode}
\usepackage{placeins}
\usepackage{hyperref}

\usepackage[preprint]{icml2026}

\usepackage{amsmath}
\usepackage{amssymb}
\usepackage{mathtools}
\usepackage{amsthm}

\usepackage[capitalize,noabbrev]{cleveref}

\theoremstyle{plain}

\theoremstyle{definition}

\theoremstyle{remark}

\usepackage[textsize=tiny]{todonotes}

\def \name{\textsc{RoboAlign}\xspace}
\icmltitlerunning{RoboAlign: Learning Test-Time Reasoning for Language-Action Alignment in Vision-Language-Action Models}

\begin{document}

\twocolumn[
  \icmltitle{RoboAlign: Learning Test-Time Reasoning for \\ Language Action Alignment in Vision Language Action Models}

  \begin{icmlauthorlist}
    \icmlauthor{Dongyoung Kim}{kaist,real}
    \icmlauthor{Sumin Park}{kaist}
    \icmlauthor{Woomin Song}{kaist}
    \icmlauthor{Seungku Kim}{kaist}
    \icmlauthor{Taeyoung Kim}{kaist}
    \icmlauthor{Huiwon Jang}{kaist,real}
    \icmlauthor{Jinwoo Shin}{kaist,real}
    \icmlauthor{Jaehyung Kim\textsuperscript{*}}{yonsei}
    \icmlauthor{Younggyo Seo\textsuperscript{*}}{berkeley}
  \end{icmlauthorlist}

  \icmlaffiliation{kaist}{KAIST}
  \icmlaffiliation{yonsei}{Yonsei University}
  \icmlaffiliation{berkeley}{UC Berkeley}
  \icmlaffiliation{real}{RLWRLD}

  \icmlcorrespondingauthor{Dongyoung Kim}{kingdy2002@kaist.ac.kr}

  \icmlkeywords{Machine Learning, ICML}

  \vskip 0.3in
]



\printAffiliationsAndNotice{* Equal advising.}  

\begin{abstract}

Improving embodied reasoning in multimodal-large-language models (MLLMs) is essential for building vision-language-action models (VLAs) on top of them to readily translate multimodal understanding into low-level actions.
Accordingly, recent work has explored enhancing embodied reasoning in MLLMs through supervision of vision-question-answering type. However, these approaches have been reported to result in unstable VLA performance, often yielding only marginal or even negative gains.
In this paper, we propose a more systematic MLLM training framework \name{} that reliably improves VLA performance.
Our key idea is to sample action tokens via zero-shot natural language reasoning and refines this reasoning using reinforcement learning (RL) to improve action accuracy. As a result, \name{} bridges the modality gap between language and low-level actions in MLLMs, and facilitate knowledge transfer from MLLM to VLA.
To validate the effectiveness of \name{}, we train VLAs by adding a diffusion-based action head on top of an MLLM backbone and evaluate them on major robotics benchmarks. 
Remarkably, by performing RL-based alignment after SFT using less than 1\% of the data, \name{} achieves performance improvements of 17.5\%, 18.9\%, and 106.6\% over SFT baselines on LIBERO, CALVIN, and real-world environments, respectively.
\end{abstract}
\section{Introduction}
\label{sec:intro}

Vision–language–action models (VLAs) have recently demonstrated remarkable success in robotics \citep{brohan2022rt,brohan2023rt,driess2023palm}.
By integrating visual perception, language understanding, and common-sense knowledge of multimodal-large-language models (MLLMs), VLAs provide a foundation for training generalizable robotic policies in real-world scenarios \citep{yang2023learning, huang2022inner, tellex2020robots, huang2022language, hu2023look}.
Specifically, policies are obtained either through discrete action token predictions by MLLMs \citep{kim2024openvla, pertsch2025fast, kim2025fine} or through continuous action prediction by external action experts that operate on latent states of MLLMs \citep{black2024pi_0, bjorck2025gr00t, team2024octo}. 
This approach leverages the extensive pretrained knowledge within MLLMs, enabling the development of generalizable policies even with a limited amount of robotics data.

However, the performance and generalization of VLAs are often limited by the underlying MLLMs, which struggle with key embodied tasks required for action generation, such as spatial reasoning \citep{tong2024cambrian, zhou2025roborefer, cheng2024spatialrgpt} and temporal reasoning \citep{ahn2022can,sermanet2024robovqa}.
To address this limitation, researchers have developed various embodied question-answering tasks designed to improve reasoning skills for robotic manipulation. 
These include tasks such as answering high-level action questions \citep{chen2025training, lynch2023interactive}, responding to spatial questions about object relationships \citep{chen2024spatialvlm, xu2025multi}, grounding points or bounding boxes in images to identify affordance-related locations \citep{yuan2024robopoint, song2025robospatial}, and predicting future visual trajectories of end-effectors \citep{ji2025robobrain, yuan2025seeing}. 
While supervised fine-tuning (SFT) has been the dominant paradigm for these tasks, recent approaches have leveraged reinforcement learning (RL) strategies (\textit{e.g.}, DeepSeek-R1; \citealt{guo2025deepseek}) to elicit stronger reasoning capabilities, leading to significant performance improvements \citep{azzolini2025cosmos,kim2025robot, song2025maniplvm, huang2025thinkact}.

\begin{figure}[t!]
    \vspace{0.25cm}
    \centering
    \includegraphics[width=0.95\linewidth]{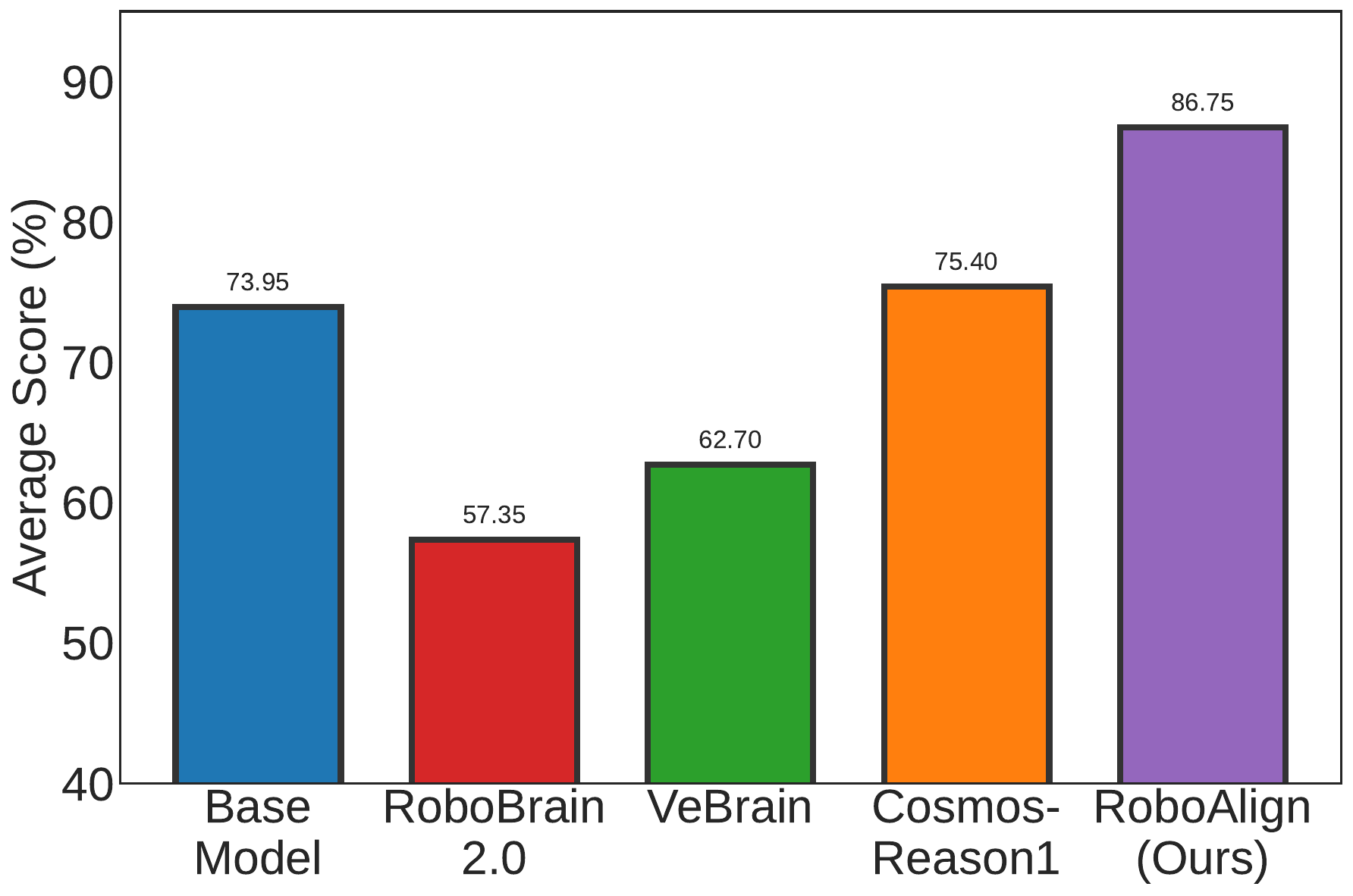} 
    \caption{\textbf{Performance on LIBERO.} 
    VLAs built upon MLLMs specialized for embodied reasoning (fine-tuned variants of Qwen2.5-VL-7B-Instruct) fail to significantly improve performance and often degrade it compared to the baseline VLA based on the original model.
    In contrast, \name{} achieves significant gains, as detailed in Section~\ref{sec:experiment}.}
    \label{fig:summary}
    \vspace{-0.2in}
\end{figure}

Despite recent successes, improvements in embodied reasoning do not consistently translate into corresponding gains in VLA performance. 
Notably, VLM4VLA~\citep{zhang2026vlm4vla} revealed that the correlation between embodied reasoning capability and VLA performance is inconsistent and highly task-dependent, sometimes even leading to performance degradation.
To further support this observation, we conducted additional experiments by training VLAs on top of open-source MLLMs specialized in embodied reasoning and observed similar trends (see Figure \ref{fig:summary}).
Surprisingly, although RoboBrain 2.0~\citep{team2025robobrain} achieved the highest reasoning scores among evaluated MLLMs and even outperformed GPT-4o~\citep{hurst2024gpt} on major benchmarks (see Table \ref{tab:mllm_result}), it yielded the lowest VLA performance (see Figure \ref{fig:summary}).
We attribute this discrepancy to the modality gap between language and low-level actions; optimizing embodied reasoning purely through language supervision does not guarantee improvements in actual action generation.

\textbf{Contribution.} Motivated by this insight, we introduce \name{}, an MLLM training framework that reliably improves VLA performance.
The core idea of \name{} lies in generating low-level action tokens as a direct outcome of embodied reasoning and evaluating reasoning quality via action accuracy; this approach allows us to directly align the reasoning capability of MLLMs to VLAs through RL-based fine-tuning.
Specifically, \name{} first applies SFT to enable the MLLM to generate low-level actions via zero-shot reasoning. It then employs GRPO~\citep{shao2024deepseekmath} to refine this reasoning process by maximizing an action-accuracy reward. This approach allows the model to explore diverse reasoning trajectories through sampling and align them toward precise action execution.

To evaluate the effectiveness of \name{}, we train MLLMs with our framework and test the performance on 
a suite of robotic benchmarks, including simulation environments such as LIBERO \citep{liu2023libero} and CALVIN \citep{mees2022calvin}, as well as real-world robot settings. 
Specifically, we attach a diffusion-based action head to the frozen MLLM backbone and fine-tune it to generate low-level actions.
Our experiments show that models trained with \name{} achieve substantial performance gains over baseline SFT-only models, with relative improvements of 17.5\% on LIBERO, 18.9\% on CALVIN, and 106.6\% in the real-world setup, while using less than 1\% additional data for the subsequent RL-based alignment stage on top of SFT.
Moreover, we find that our approach is more effective than other alignment approaches such as high-level action prediction (13.1\% v.s. 17.5\%) or point trajectory prediction  (15.2\% v.s. 17.5\%) on the LIBERO, respectively.

Furthermore, to examine if \name{} also improves embodied reasoning capabilities of MLLMs, we evaluated \name{} on a diverse set of benchmarks for general image understanding~\citep{chen2024we}, spatial reasoning~\citep{song2025robospatial,yuan2024robopoint,fu2024blink}, and embodied reasoning for robotics~\citep{kim2025robot}.
On the embodied reasoning tasks, \name{} achieve state-of-the-art performance, outperforming not only commercial general-purpose models such as GPT-4o \citep{openai2024gpt4o}, but also specialized embodied MLLMs, such as RoboBrain2.0 \citep{team2025robobrain}.
Notably, this is accomplished while preserving the model's performance on general image understanding.
This result shows that our RL-based alignment enhances the general capabilities of MLLMs,
in contrast to SFT-based alignment methods such as ECoT~\citep{zawalski2024robotic}, which often degrades performance on these embodied tasks.
\vspace{-0.2cm}
\section{Related Work}
\label{sec:relatedwork}

\textbf{Multimodal-large-language models for robot control.} Efforts to leverage the visual processing capabilities, commonsense, and world knowledge of multimodal-large-language models (MLLMs) for robot policy decision have shown consistent success. In particular, MLLMs have demonstrated strong performance in high-level action planning. Concretely, prior work has explored generating predefined atomic action skills to directly control robots~\citep{liang2023code,tellex2020robots,luo2025visual}, or producing high-level actions and plans that condition subsequent low-level actions~\citep{driess2023palm,yang2023learning,huang2022inner,huang2022language,hu2023look}. These approaches have been further extended toward more precise action generation, either by enabling MLLMs to produce policies in an end-to-end manner~\citep{kim2024openvla,pertsch2025fast,kim2025fine} or by training action experts that consume latent states instead of language outputs~\citep{team2024octo,li2023vision,shentu2024llms,black2024pi_0,bjorck2025gr00t,nvidia2025gr00t}. We investigate how to better align MLLMs with low-level actions to enhance such robot control performance.

\textbf{Multimodal-large-language model for embodied reasoning.} With the increasing application of MLLMs to embodied environments such as robot manipulation, their capabilities for tasks requiring spatial and temporal reasoning have been enhanced. For spatial reasoning, prior work has enhanced 3D scene understanding by leveraging VQA data to train models that convert information from 2D and 3D vision inputs~\citep{chen2024spatialvlm,ray2024sat,zhou2025roborefer,wu2025spatial}. To further improve performance in specific robotic tasks, some approaches have trained models to predict bounding boxes or points associated with affordances and manipulation-relevant spatial cues~\citep{yuan2024robopoint,song2025robospatial,lu2023vl,ji2025robobrain}. For temporal reasoning, researchers have extracted high-level actions~\citep{chen2025training, lynch2023interactive, chen2025training,huang2024egoexolearn,chen2023egoplan}, 2D point trajectories of object movement from egocentric videos of humans or robots to construct VQA~\citep{huang2025thinkact,yang2025magma,ranasinghe2024understanding,zheng2024tracevla,lee2025molmoact}. Nevertheless, these approaches primarily provide indirect supervision signals and do not directly optimize low-level action generation.

\textbf{Encouraging reasoning through reinforcement learning.} Chain-of-Thought (CoT) prompting~\citep{wang2022self,yao2023tree,kim2023cot,wei2022chain} has been widely applied to both LLMs and MLLMs in zero-shot, few-shot, and supervised fine-tuning (SFT) settings~\citep{muennighoff2025s1}, effectively improving answer quality. Recently, DeepSeek-R1~\citep{guo2025deepseek} proposed a training approach specialized for CoT, in which reasoning is explicitly enforced during the response process, and the entire reasoning trace is optimized using the reinforcement learning algorithm with rewards derived from the final answer. This training paradigm has demonstrated that, compared to SFT, models can achieve stronger performance and generalization across diverse domains, including mathematics~\citep{zeng2025simplerl,yu2025dapo}, agents~\citep{lu2025ui,jin2025search}, visions~\citep{shen2025vlm, huang2025vision,huang2025vision}, and embodied reasoning~\citep{kim2025robot,song2025maniplvm,huang2025thinkact,yuan2025seeing,yuan2025embodied}, while requiring significantly less data, in some cases even a single example~\citep{wang2025reinforcement}.
In this work, we introduce a reinforcement learning scheme based on low-level action prediction, aligning the MLLM’s representations more directly with robot control.

\vspace{-0.2cm}

\section{Preliminaries}
\label{sec:preliminaries}

\textbf{FAST action tokenization.} 
We adopt FAST tokenization~\citep{pertsch2025fast} to integrate low-level actions into multimodal-large-language models (MLLMs), as it has been shown to be effective not only for end-to-end policy learning but also for representation learning~\citep{black2025pi0,driess2025knowledge}.
Each action is defined as a $D$-dimensional vector representing the 
end-effector's state, which consists of its Cartesian position $(x, y, z)$, orientation $(\text{roll}, \text{pitch}, \text{yaw})$, and gripper state (Open/Close). 
An action sequence over a horizon of $H$ timesteps forms a chunk, $\mathbf{a}_{1:H} = [[a_{1,1}, a_{1,2}, \ldots, a_{1,D}], \ldots, [a_{H,1}, a_{H,2}, \ldots, a_{H,D}]]$.
To improve compactness, FAST tokenization transforms the action chunk $\mathbf{a}_{1:H}$ into the frequency domain using a discrete cosine transform (DCT; \citealt{ahmed2006discrete}).
The resulting DCT coefficients are quantized and flattened into a sequence.
This sequence is then compressed into discrete tokens using byte-pair encoding (BPE; \citealt{gage1994new}), resulting in $T_{k} = \text{FAST}(\mathbf{a}_{1:H})$, where each token is mapped to one of $2K$ special tokens added to the MLLM's vocabulary for training and generation.

\textbf{Encouraging reasoning with GRPO.} 
To encourage explicit reasoning, we train the model to generate intermediate thoughts enclosed within \texttt{<think>...</think>} before producing a final answer. 
Training is conducted with Group Relative Policy Optimization (GRPO;~\citealt{shao2024deepseekmath}), where the policy is optimized jointly for format correctness and answer accuracy.
Specifically, let the current policy be denoted as $\pi_{\theta_\text{old}}$.
For a given query $q \sim P(Q)$, we sample $G$ responses $[o_1, \ldots, o_G] \sim \pi_{\theta_\text{old}}(q)$.
Each response is evaluated by a pre-defined reward model $R(q, o_i)$, which assigns a reward $r_i$ based on format and answer accuracy.
We then compute an advantage by normalizing the reward using the standard deviation, $A_i = \frac{r_i - \text{mean}(\mathbf{r})}{\text{std}(\mathbf{r})}$, and define the importance sampling ratio as $\rho_i(\theta) = \frac{\pi_{\theta}(o_i | q)}{\pi_{\theta_\text{old}}(o_i | q)}$.
For each query $q \sim P(Q)$, we sample $G$ responses $\{o_i\}_{i=1}^G$ from the old policy $\pi_{\theta_\text{old}}$.
GRPO optimizes the policy by maximizing these advantages while applying a KL penalty against a reference policy:
\begin{equation}
\resizebox{0.9\columnwidth}{!}{$
\begin{aligned}
\mathbb{J}_{\tt{GRPO}}(\theta) = \mathbb{E}\Big[&\frac{1}{G}\sum_{i=1}^{G}\min\big(\rho_i A_i, \text{clip}(\rho_i, 1{-}\varepsilon, 1{+}\varepsilon)A_i\big) 
- \beta \mathbb{D}_{\text{KL}}(\pi_{\theta} \| \pi_{\text{ref}})\Big],
\end{aligned}
$}
\label{eq:grpo}
\end{equation}
where $\varepsilon$ and $\beta$ are hyperparameters for clipping and KL penalty.

\vspace{-0.2cm}

\section{\name{}: Align Embodied Reasoning with Low-level Actions}
\label{sec:method} 
\begin{figure*}[tb]
    \centering
    \includegraphics[width=\textwidth, height=0.5\textwidth]{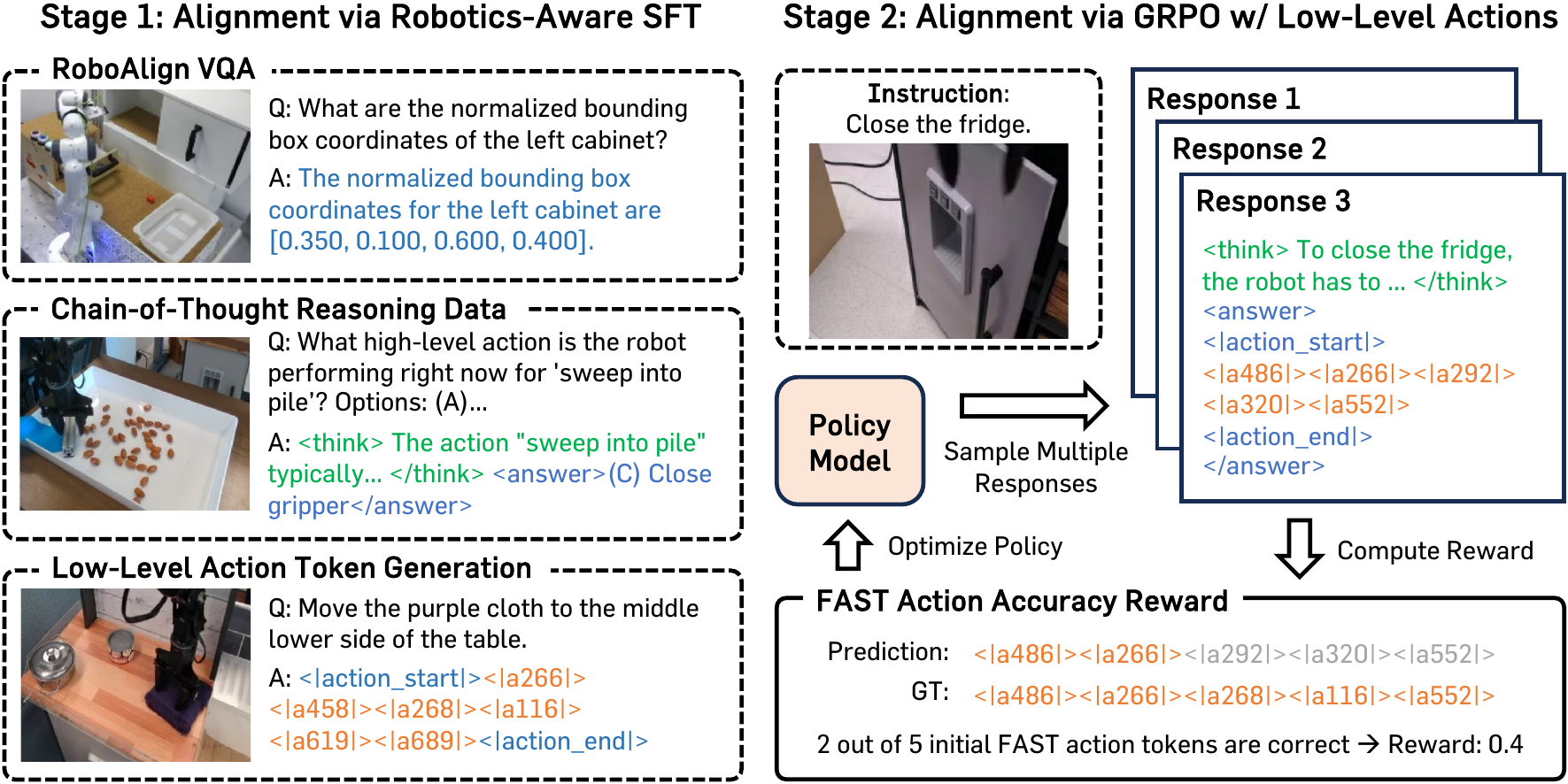}
    \caption{\textbf{Overview of \name{} framework.} \name{} directly aligns MLLM representations with low-level action generation using reasoning-incentivized reinforcement learning~\citep{guo2025deepseek}. The framework consists of two stages: (i) Stage 1 integrates embodied reasoning, zero-shot reasoning, and FAST-tokenized low-level action generation
    via supervised fine-tuning, and (ii) Stage 2 optimizes responses through reinforcement learning to improve token-level action accuracy and better alignment. The resulting model serves as an MLLM tailored for effective VLA training. }
    \label{figure:main_method}
\end{figure*}
\begin{table*}[t]
\centering
\caption{\textbf{Example of \name{} response.} Incorporating reasoning data during SFT effectively transfers zero-shot reasoning ability to FAST token generation process. Without such data, the model produces only minimal reasoning which reduces the diversity of reasoning samples and hinders RL.}
\vspace{0.05in}
\begin{adjustbox}{max width=1.0\textwidth}
\begin{tabular}{m{0.15\textwidth} m{0.85\textwidth}}
\toprule
\multirow{6}{*}{\includegraphics[width=0.16\textwidth]{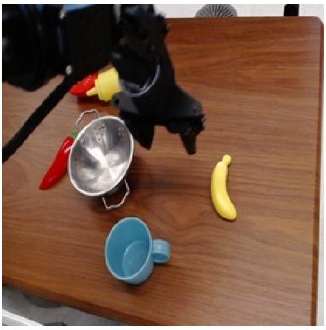}} & 
\textbf{Question:} Your current task is `pick up cup from the table'. Output the robot's actions to perform this task through FAST tokens. \\
& \\[-0.3em]
& \textbf{Model Answer (Trained w/ reasoning datasets):} <think>To pick up the cup from the table, the robot needs to move its gripper towards the cup, position it correctly, close the gripper to secure the cup, and then lift it up. Given the current state of the gripper being open and positioned above the cup, the immediate next action would be to move down towards the cup to prepare for grasping.</think><answer><|action\_start|><|action\_266|><|action\_299|> ... \\
& \\[-0.3em]
& \textbf{Model Answer (Trained w/o reasoning datasets):} <think>Go to the cup.</think> <answer><|action\_start|><|action\_266|><|action\_299|>... \\
\bottomrule
\end{tabular}
\end{adjustbox}
\label{tab:reasoning_sampling}
\end{table*}

In this section, we introduce \name{}, a training framework that directly aligns multimodal-large-language models (MLLMs) with low-level actions through reinforcement learning (RL). 
While doing so, \name{} is designed to preserve the general capabilities of MLLMs and simultaneously enhance embodied reasoning ability. 
A key challenge, however, is that off-the-shelf MLLMs cannot generate specialized low-level actions (\textit{e.g.}, FAST tokens) in a zero-shot manner, making RL inapplicable.
To address this, we introduce a two-stage training pipeline. 
First, we apply supervised fine-tuning (SFT) to equip the model with the initial ability to predict FAST tokens through zero-shot reasoning, while preserving the general abilities of MLLMs and enhancing embodied reasoning. 
Second, building on this ability, we apply RL on this SFT model to further strengthen embodied reasoning and improve FAST token prediction accuracy.
The overall process is illustrated in Figure~\ref{figure:main_method}.

\subsection{Stage 1: Integrating Low-level Action with MLLM using SFT}\label{sec:4.2}
The primary objective of this SFT stage is to equip the MLLM with the ability to generate FAST action tokens, which is a prerequisite for the subsequent RL stage, while simultaneously preserving its general vision-language capabilities and enhancing its embodied reasoning skills.
To achieve this, we curate a data mixture from four sources: (i) a variety of open-source SFT datasets for embodied reasoning and general understanding, (ii) our custom \name{} VQA dataset for robotic embodied reasoning, (iii) specialized reasoning datasets designed to improve zero-shot reasoning of MLLMs, and (iv) robotic dataset with FAST tokens. 
We describe the process for building our custom datasets in this section, with full details for all data sources and configurations available in Appendix~\ref{appendix:detail}. 

\textbf{\name{} VQA.}  
While existing VQA datasets are useful for general embodied reasoning, high-quality VQA specifically grounded in robotic information remains limited. 
For example, datasets such as ShareRobot~\citep{ji2025robobrain} and RoboVQA~\citep{sermanet2024robovqa} use robot imagery but focus on high-level QA tasks, lacking the fine-grained, spatial-temporal information needed for low-level control. 
To address this gap, we develop a data generation pipeline that feeds robot images and associated metadata, \textit{e.g.}, bounding boxes, end-effector states, and both high and low-level actions, into a powerful large model, \textit{i.e.,} \texttt{gemini-2.5 pro}~\citep{googledeepMind2025geminiUpdate}.
The model then automatically generates a diverse set of high-quality VQA, captioning, and grounding QA pairs.

\textbf{Reasoning dataset with zero-shot CoT.}  
To preserve the MLLM's zero-shot reasoning ability during SFT and transfer it to the action generation process, we incorporate a specialized reasoning dataset into our training mixture.
This dataset is created by distilling outputs from a reasoning model that is trained with GRPO to generate step-by-step reasoning. 
Specifically, we first train the reasoning model on spatial and robot-related embodied MCQAs for distillation, following \citet{kim2025robot}. 
For each prompt, we sample multiple reasoning trajectories from this model.
These outputs are then filtered using a combination of rule-based rewards and correctness checks.
Table~\ref{tab:reasoning_sampling} shows that including this specialized reasoning data during SFT enables the effective transfer of reasoning ability to FAST token generation, while the absence of such data results in limited zero-shot reasoning.

\textbf{FAST token generation dataset.}  
To enable FAST token prediction, we first extend the MLLM's vocabulary by adding two special marker tokens \texttt{<ACTION\_START>}, \texttt{<ACTION\_END>} and 2K FAST tokens. 
The training data is then constructed from the BridgeV2 dataset \citep{walke2023bridgedata}
in a QA format.
Each sample pairs a robot image with a fixed instruction, where the ground-truth answer is the corresponding sequence of FAST tokens.
The resulting data mixture, consisting of our custom and open-source datasets, is used to fine-tune the MLLM with SFT, providing a strong foundation for subsequent RL training stage.

\subsection{Stage 2: Aligning Embodied Reasoning with Low-level Action via RL}\label{sec:4.4}

In the second stage, we use RL to directly align the MLLM with low-level actions, \textit{i.e.,} FAST tokens, further refining the model to be better suitable for VLA adaptation.
Specifically, we optimize the model's embodied reasoning process to directly improve the accuracy of FAST action token generation.
To create the data for this stage,
we adapt the FAST token dataset from Stage 1.
In particular, each sample's input instruction is augmented with a prompt that requires explicit reasoning within \texttt{<think>...</think>} tags before producing the FAST token sequence.

We define the reward as the arithmetic mean of two components: a format reward $r_f \in \{0,1\}$ indicating whether the output correctly adheres to the required reasoning format, and an accuracy reward $r_a \in [0,1]$ measuring FAST token prediction accuracy. 
In particular, the accuracy reward $r_{a}$ is computed by measuring the prefix similarity between the generated action token sequence $T^{\text{gen}}_{1:n}$ and the target sequence $T^{\text{target}}_{1:m}$, normalized by the target length:
\begin{equation}
r_a = \frac{1}{m}\max \{ i \in \{1,\dots,m\} : T^{\text{gen}}_{1:i} = T^{\text{target}}_{1:i} \}.
\label{eq:accuracy_fomulation}
\end{equation}
The final reward is given by $r = (r_f + r_a)/2$. 
This formulation encourages the model to generate both correctly formatted and accurate FAST token sequences. 
Building on the constructed training dataset and reward function, we then apply GRPO \citep{shao2024deepseekmath} to further optimize the MLLM. 
\vspace{-0.04in}
\section{Experiment}\label{sec:experiment}
\vspace{-0.04in}

\begin{figure}[t]
    \centering
    \includegraphics[width=\columnwidth]{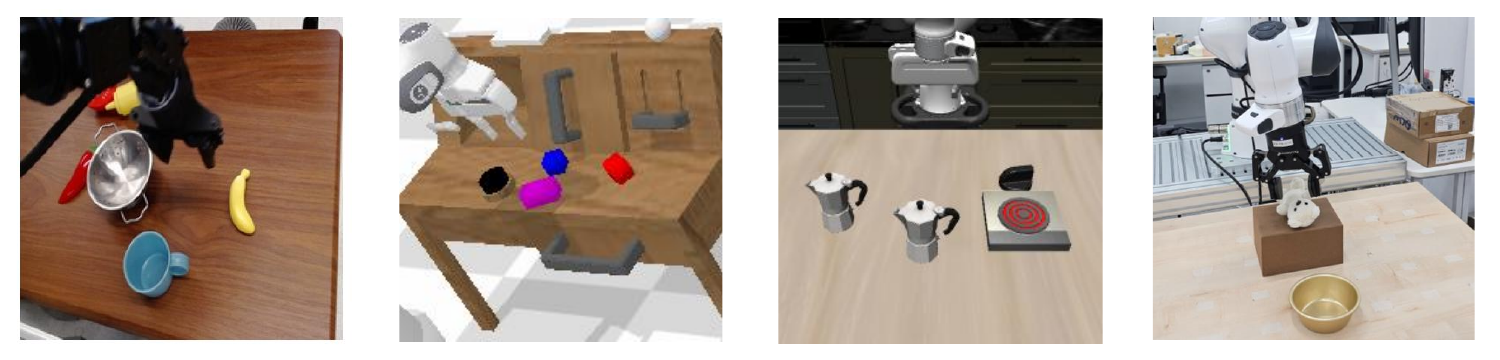} 
    \caption{\textbf{Examples of observations.} Visual inputs for training and evaluation (from left to right): BridgeV2 for FAST token training, CALVIN, LIBERO benchmark, and the real-robot.}
    \label{fig:robot_images}
    \vspace{-0.2in}
\end{figure}
\begin{figure}[t]
    \centering
    \includegraphics[width=0.95\linewidth]{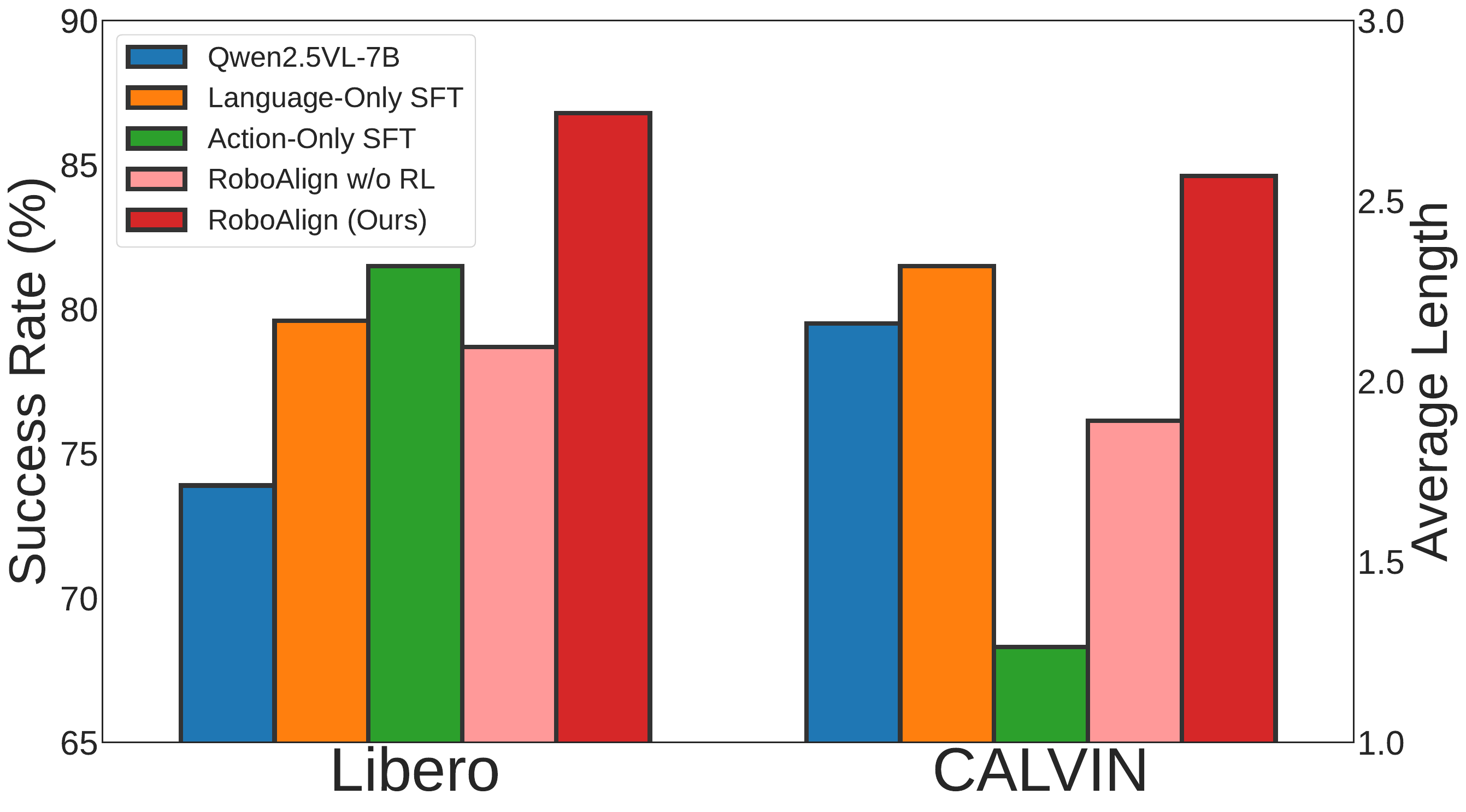} 
    \caption{\textbf{Summary of VLA performance.} Comparison of VLA performance across different MLLM training methods on LIBERO and CALVIN. 
    \name{} achieves the highest gains.
    }
    \label{fig:result_sumamry}
    \vspace{-0.5cm}
\end{figure}

\begin{table*}[t]
\caption{\textbf{LIBERO} success rates (\%) for VLAs built upon MLLMs that were fine-tuned with various methods, evaluated over 500 trials per category.
Each model is evaluated by training a newly-initialized, diffusion-based action head on the LIBERO dataset while the MLLM backbone remains frozen. \name{} shows particularly large improvements in the Long and Goal categories compared to other training methods.}
\centering\small
\begin{adjustbox}{max width=0.95\textwidth}
    \begin{tabular}{l  c | cccc | c}
        \toprule
        Method & Fine-tuning Dataset Size & Spatial & Object & Goal & Long & Avg. \\
        \midrule
        \midrule
        \multicolumn{7}{l}{\textit{Reference VLA Performance}} \\
        \midrule
        Diffusion Policy & - & 82.6 & 84.7 & 82.1 & 57.6 & 76.8 \\
        Octo \citep{team2024octo} & - & 78.9 & 85.7 & 84.6 & 51.1 & 75.1 \\
        OpenVLA \citep{kim2024openvla} & - & 84.9 & 88.4 & 79.2 & 53.7 & 76.5 \\
        TraceVLA \citep{zheng2024tracevla} & - & 84.9 & 85.2 & 75.1 & 54.1 & 74.8 \\
        CoT-VLA \citep{zhao2025cot} & - & 87.5 & 91.6 & 87.6 & 69.0 & 83.9 \\
        ThinkAct \citep{huang2025thinkact} & - & 88.3 & 91.4 & 87.1 & 70.9 & 84.4 \\
        \midrule
        \midrule
        \multicolumn{7}{l}{\textit{Our Setup (Baselines and Ours)}} \\
        \midrule
        Qwen2.5VL-7B-Ins~\citep{bai2025qwen2} & - & 95.2 & 95.0 & 42.4 & 63.2 & 73.9 \\
        w/ Language-Only SFT & 1.88M  & 91.0 & 94.4 & 67.8 & 65.0 & 79.6 \\
        w/ Action-Only SFT & 1.88M   & 89.8 & 95.8 & 82.8 & 57.6 & \underline{81.5} \\
        \rowcolor{gray!20} w/ \name{} w/o RL & 2.28M & 92.8 & 97.4 & 59.0 & 65.6 & 78.7 \\
        \rowcolor{gray!20} w/ \name{} (Ours) & 2.28M+12.8K & 93.8 & 96.0 & 87.2 & 70.0 & \textbf{86.8} \\
        \bottomrule
    \end{tabular}
\end{adjustbox}
\label{tab:libero}
\end{table*}
\begin{table*}[t]
\caption{\textbf{CALVIN ABC$\to$D} success rates (\%) for VLAs built upon MLLMs that were fine-tuned with various methods, evaluated over 1000 trials.
Each model is evaluated by training a newly-initialized, diffusion-based action head on the CALVIN dataset while the MLLM backbone remains frozen. While all baselines show drops in task completions of length 4 and 5, \name{} consistently improves performance across all sequence. 
}
\centering\small
\begin{adjustbox}{max width=0.95\textwidth}
\begin{tabular}{l|c|ccccc|c}
\toprule
\multirow{2}{*}{Method} & Fine-tuning & \multicolumn{5}{c|}{Task completed in a row (\%) $\uparrow$} & Succ. Len.   \\
                        & Dataset Size & 1          & 2          & 3         & 4         & 5         & (Avg)  \\
\midrule
Qwen2.5VL-7B-Ins~\citep{bai2025qwen2}            & -      & 77.8       & 55.0       & 38.6      & 26.6      & 18.1      & 2.16 \\
w/ Language-Only SFT & 1.88M & 87.4       & 62.2       & 41.9      & 25.2      & 15.3      & \underline{2.32} \\
w/ Action-Only SFT & 1.88M  & 66.1       & 34.7       & 15.3      & \phantom{0}7.1       & \phantom{0}3.2       & 1.26 \\
\rowcolor{gray!20} w/ \name{} w/o RL             & 2.28M      & 74.6       & 49.6       & 31.5      & 21.2      & 12.2      & 1.89 \\
\rowcolor{gray!20} w/ \name{} (Ours)         & 2.28M+12.8K      & 87.6       & 67.2       & 47.1      & 32.8      & 22.2      & \textbf{2.57}
\\ \bottomrule
    \end{tabular}\label{tab:calvin} 
\end{adjustbox}
\vspace{-0.2cm}
\end{table*}
\begin{table*}[t]
\caption{\textbf{Real robot} success rates (\%) for VLAs built upon MLLMs that were fine-tuned with various methods, evaluated over 96 trials per task.
Each model is evaluated by training a newly-initialized, diffusion-based action head on the real-world robotic dataset while the MLLM backbone remains frozen. We find that \name{} is also effective in real-world settings.
}
\centering\small
\begin{adjustbox}{max width=0.95\textwidth}
    \begin{tabular}{l | cccc | c}
        \toprule
        Method  & \shortstack{Box to \\ bowl} & \shortstack{Box to \\ plate} & \shortstack{Basket to \\ bowl} & \shortstack{Plate to \\ basket} & Avg. \\
        \midrule
        Qwen2.5VL-7B-Ins~\citep{bai2025qwen2}  & 16.7 &	70.8 &	20.8 &	20.8 &	32.3 \\ 
        \rowcolor{gray!20} w/ \name{} w/o RL  & 87.5 & 	58.3 & 	37.5 & 	37.5 & 	\underline{55.2}\\
        \rowcolor{gray!20} w/ \name{} (Ours) & 87.5 & 	58.3 	 & 70.8 &	50.0  &	\textbf{66.7} \\
        \bottomrule
    \end{tabular}\label{tab:real_exp} 
\end{adjustbox}
\end{table*}
In this section, we design experiments to answer the following research questions:
\begin{itemize}[leftmargin=5.5mm,topsep=0pt,itemsep=0pt]
    \item[$\circ$] Does training MLLMs with \name{} consistently improve VLA performance across various scenarios? (See Table~\ref{tab:libero}, \ref{tab:calvin}, \ref{tab:real_exp}, \ref{tab:qwen3_libero} and Figure~\ref{fig:result_sumamry})
    \item[$\circ$] Is \name{} more effective than alternative MLLMs training methods? (See Table~\ref{tab:change_aling_libero}, \ref{tab:sft_vs_rl_libero})
    \item[$\circ$] How does \name{} contribute to VLA performance improvements? (See Table~\ref{tab:knn})
\end{itemize}
\vspace{-0.03in}
\subsection{Experimental Setup}
\vspace{-0.03in}

\textbf{Experiment design.} To evaluate how different MLLM training methods affect VLA performance, we convert MLLMs trained with various algorithms into VLAs using an identical robot dataset and a unified VLA conversion pipeline, and then evaluate their robot control performance.
Our VLA conversion pipeline is built upon the well-established VLA training framework Gr00t-N1.5~\citep{nvidia2025gr00t}. Concretely, we adapt diffusion-based action head on top of an MLLM backbone and train newly-initialized diffusion-based action head on robot datasets while keeping the MLLM backbone frozen. For each benchmark, we train VLAs from scratch using the training data provided by the benchmark.

\textbf{MLLMs training data.} For supervised fine-tuning (SFT), we prepare a diverse set of datasets covering both general MLLM capability and FAST token prediction. 
In total, 1.88M samples are used for MLLM-related tasks. 
For FAST token prediction, we use the subset of BridgeV2~\citep{walke2023bridgedata} dataset (400K samples), yielding 2.28M samples overall. 
For reinforcement learning (RL), we further use a 12.8K subset of the BridgeV2 FAST token prediction data.
More details are provided in Appendix~\ref{appendix:detail}.

\textbf{Baseline models.} 
To validate the effectiveness of \name{}, we prepare two baselines: (i) a model trained only on MLLM data and (ii) a model trained only on FAST token prediction using the full BridgeV2 dataset (1.88M samples). 
Both are trained for one epoch following the same SFT train schema as in \name{}.

\textbf{Benchmarks.} We evaluate VLA performance in LIBERO~\citep{liu2023libero}, CALVIN~\citep{mees2022calvin} and Real robot (see Figure~\ref{fig:robot_images} for the examples). 
\begin{itemize}[leftmargin=5.5mm,topsep=0pt,itemsep=0pt]
    \item[$\circ$] \textbf{LIBERO:} This benchmark uses a Franka Panda Arm to perform manipulation tasks grouped into four categories: spatial, object, goal, and long-horizon. 
    Each category consists of 10 tasks. 
    Training uses the provided dataset covering all tasks, and evaluation runs 50 trials per task (500 trials per category). 

    \vspace{-0.2cm}
    
    \item[$\circ$] \textbf{CALVIN:} 
    This benchmark also employs a Franka Panda Arm and consists of 34 distinct tasks. 
    Training uses data collected from environments A, B, and C for 100K steps, after which zero-shot evaluation is performed in a novel environment D. 
    Performance is measured by the success rate of executing five consecutive instruction chains, with a total of 1,000 chains evaluated. 
\end{itemize}

\textbf{Implementation details.} 
We train our models based on Qwen2.5VL-7B-Ins~\citep{bai2025qwen2}. 
For SFT, we follow the official Qwen2.5VL training repository. 
The vision encoder is frozen, and we use a cosine scheduler with a learning rate of $2 \times 10^{-5}$, a warmup ratio of 0.03 and training for 1 epoch.
For RL, we use the EasyR1 repository\footnote{https://github.com/hiyouga/EasyR1}, training all parameters from scratch with a rollout batch size of 512, update batch size of 128, and 5 samples per prompt. We apply a constant learning rate of $1 \times 10^{-6}$ and train for one epoch. For VLA training, see detail in Appendix~\ref{appendix:detail}. 
\subsection{Main Results}\label{sec:main} 
\vspace{-0.6cm} 

\

As shown in Tables \ref{tab:libero}, \ref{tab:calvin}, MLLMs trained with \name{}, which combines SFT and RL, achieve the highest performance across all simulations. 
The SFT stage alone yields moderate improvements, suggesting that most of the performance gain comes from the RL stage, despite using less than 1\% of the data (2.28M vs 12.8K) used in the SFT stage.
In particular, \name{} demonstrates a significant increase in success rates on long-horizon tasks, which are more intricate and complex than other types of tasks.
For example, in CALVIN (Table~\ref{tab:calvin}), \name{} achieves the highest task completions of length-5 success rate (18.1\% $\rightarrow$ 22.2\%), whereas all other training methods  exhibit a decline in performance.
Similarly, in LIBERO (Table~\ref{tab:libero}), the \textit{Long} category improves to 70\% with \name{}, compared to only $\sim$2\% gains from other methods. 

Another notable finding is in the \textit{Goal} category of LIBERO, which requires handling different instructions in the same environment. 
Here, \name{} improves performance dramatically from 42.4\% to 87.2\%. 
However, models trained only with MLLM data show limited improvements. Specifically, in CALVIN they achieve higher success in task completions of length-1 (77.8\% $\rightarrow$ 87.4\%) but experience a drop in task completions of length-5 performance (18.1\% $\rightarrow$ 15.3\%). Similarly, in LIBERO they improve in the \textit{Goal} category (42.4\% $\rightarrow$ 67.8\%) but yield only marginal gains in the \textit{Long} category (63.2\% $\rightarrow$ 65.6\%).
These results indicate that embodied reasoning abilities learned through language can enhance performance on relatively simple tasks, but offer limited improvements on more complex and demanding tasks.
When trained only with VLA data, we observe large in-domain gains, particularly in LIBERO’s \textit{Goal} category (42.4\% $\rightarrow$ 82.8). 
However, performance drops significantly on long-horizon tasks in both CALVIN and LIBERO. 
We hypothesize that while FAST token training strengthens alignment between instructions and low-level actions in-domain, it also induces forgetting of general MLLM capabilities, leading to reduced zero-shot generalization.

\subsection{Ablation Study and Analyses}\label{sec:5.3}

\textbf{Real robot experiments.}  To examine whether the improvements of \name{} on VLA performance extend beyond simulation to real-robot settings, we conduct experiments using a Franka Research 3 robot arm across four distinct pick-and-place tasks. Each task involves moving a different object (teddy bear, box, cup, sponge). Training is performed with 60 demonstrations per task, and evaluation consists of 24 trials per object, totaling 96 trials per task. The VLA setup follows the same configuration as in the main experiments, with each task trained for 30K steps. As shown in Table~\ref{tab:real_exp}, \name{} consistently improves performance even in real-robot settings.

\begin{table}[t]
\caption{\textbf{Compatibility with different models.} 
We apply \name{} to a different MLLM backbone (Qwen3VL-8B-Ins) to validate its generalizability. We report success rates (\%) on the LIBERO benchmark, averaged over 500 trials per category. \name{} consistently improves overall performance, with particularly significant gains in the Long category.
}
\centering\small
\begin{adjustbox}{max width=\columnwidth}
    \begin{tabular}{l | cccc | c}
        \toprule
        Method & Spatial & Object & Goal & Long & Avg. \\
        \midrule
            Qwen3VL-8B-Ins~\citep{qwen2025qwen3vl}              & 94.2  & 96.4 & 90.0  & 60.0  & 85.2 \\
            \rowcolor{gray!20} w/ \name{} w/o RL          & 96.2 & 97.4 & 93.3 & 71.0 &  89.5 \\
            \rowcolor{gray!20} w/ \name{} (Ours)     & 95.6 & 99.6 & 95.2 & 78.6 & \textbf{92.5} \\
        \bottomrule
    \end{tabular}\label{tab:qwen3_libero} 
\end{adjustbox}
\vspace{-0.3cm}
\end{table}
\textbf{Compatibility with different models.} To assess whether \name{} generalizes to other architectures, we conducted experiments using another MLLM backbone, Qwen3-VL-8B-Ins. For the MLLM training phase, we utilized 5K samples for RL, while maintaining all other training setup. After training, all models are converted into VLAs and evaluated on the LIBERO simulation environments. As shown in Table~\ref{tab:qwen3_libero}, we observed an overall performance increase, with particularly significant gains in the Long category. This trend is consistent with the results observed in Table~\ref{tab:libero}. These results demonstrate that \name{} effectively generalizes across different MLLM architectures.

\begin{table}[t]
\caption{\textbf{Impact of alignment strategies on VLA.} 
We compare RL alignment with different targets and evaluate on LIBERO, reporting success rates (\%): Language-based RL (high-level language actions), Visual-based RL (2D end-effector trajectories), and Action-based RL (Ours, low-level actions). \name{} consistently improves performance and uniquely enhances long-horizon tasks.
}
\centering\small
\begin{adjustbox}{max width=\columnwidth}
    \begin{tabular}{l | cccc | c}
        \toprule
        Method  & Spatial & Object & Goal & Long & Avg. \\
        \midrule
            \rowcolor{gray!20} \name{} w/o RL & 92.8 & 97.4 & 59.0 & 65.6 & 78.7  \\
            w/  Language-base RL &91.6 & 94.6 & 90.0 & 58.2 & 83.6 \\
            w/  Visual-base RL &92.4 & 95.6 & 87.8 & 64.6 & 85.1\\
            \rowcolor{gray!20} w/ Action-base RL (Ours) &  93.8 & 96.0 & 87.2 & 70.0 & \textbf{86.8} \\
        \bottomrule
    \end{tabular}\label{tab:change_aling_libero} 
\end{adjustbox}
\vspace{-0.1cm}
\end{table} 
\textbf{Comparison with embodied alignment strategies.} 
To evaluate the effectiveness of the low-level action–based RL alignment in \name{}, we compare it with two commonly used embodied MLLM training tasks that also employ RL: (i) predicting high-level actions expressed in language descriptions and (ii) predicting 2D visual trajectories of the end effector.
As shown in Table~\ref{tab:change_aling_libero}, \name{} achieves the largest performance improvement. 
In contrast, the alternative methods show notable gains in the LIBERO \textit{Goal} category but remain limited on long-horizon tasks. 
This trend is consistent with our main results in Section~\ref{sec:5.3} and further demonstrates the advantage of direct alignment with low-level actions (See Appendix~\ref{appendix:alignment_setup} for details).

\begin{table}[t]
\caption{\textbf{Comparison with SFT-based alignment.} 
We compare our RL-based alignment against an SFT-based baseline that jointly trains reasoning and low-level action. Both methods are fine-tuned from \name{} SFT model and evaluated on the LIBERO benchmark. While the SFT-based baseline degrades performance, \name{} achieves significant improvements.
}
\centering\small
\begin{adjustbox}{max width=\columnwidth}
    \begin{tabular}{l | cccc | c}
        \toprule
        Method  & Spatial & Object & Goal & Long & Avg. \\
        \midrule
        \rowcolor{gray!20} \name{} (SFT)  & 92.8 & 97.4 & 59.0 & 65.6 & 78.7  \\
        w/ SFT-based Alignment (ECoT) & 84.6 &	90.8 &	49.6 &	45.6 &	67.7 \\
        \rowcolor{gray!20} w/ RL-based Alignment (Ours) &  93.8 & 96.0 & 87.2 & 70.0 & \textbf{86.8} \\
        \bottomrule
    \end{tabular}\label{tab:sft_vs_rl_libero} 
\end{adjustbox}
\vspace{-0.2cm}
\end{table}
\textbf{Comparison with SFT-based alignment.} 
We further compare RL-based alignment in \name{} against SFT-based alignment using ECoT~\citep{zawalski2024robotic}, which jointly trains reasoning and low-level actions through SFT.
As shown in Table~\ref{tab:sft_vs_rl_libero}, the SFT-based approach even reduces performance compared to RL. 
We attribute this to the limited generalization of SFT, where knowledge aligned on BridgeV2 transfers poorly to LIBERO, as well as to forgetting effects introduced during SFT.
Consistently, when evaluated on general MLLM benchmarks, the ECoT-trained model shows a degradation in performance, confirming the limitations of SFT-based alignment (See Appendix~\ref{appendix:alignment_setup} for detailed experimental setup).

\begin{table}[t]
\caption{\textbf{K-Nearest neighbor accuracy.} We measure how accurately MLLM representations can predict underlying states using KNN classification on 20 trajectories from a LIBERO task.}
\centering\small
\begin{adjustbox}{max width=0.7\textwidth}
    \begin{tabular}{l | c}
        \toprule
        Method & Acc. (\%) \\
        \midrule
            Qwen3VL-8B-Ins~\citep{qwen2025qwen3vl}              & 39.06 \\
            \rowcolor{gray!20} w/ \name{} w/o RL                                     & 43.23 \\
            \rowcolor{gray!20} w/ \name{} (Ours)                                & \textbf{69.79} \\
        \bottomrule
    \end{tabular}
\end{adjustbox}
\vspace{-0.7cm}
\label{tab:knn}
\end{table}

\textbf{Representation analysis.} To understand how \name{} improves VLA performance, we analyze the internal representations of MLLMs. We hypothesize that \name{} explicitly aligning low-level actions enables the model to learn more discriminative and fine-grained features for action generation. To test this hypothesis, we design a KNN classification experiment to evaluate whether the MLLM representation alone, without access to robot state information, can distinguish between similar underlying states. As shown in Table~\ref{tab:knn}, \name{} (SFT+RL) produces substantially more discriminative representations than both baselines, improving KNN accuracy from $39.06\%$ to $69.79\%$. 
This result indicates that the RL alignment stage significantly sharpens the model's ability to encode fine-grained state information. 
Distinct representations help to generate accurate actions, and these results help to understand the mechanism of \name{}'s performance improvement (See Appendix~\ref{appendix:alignment_setup} for detailed setups).

\textbf{Performance on MLLM Benchmarks}
To examine whether \name{} enhances not only VLA alignment but also embodied reasoning and general MLLM capabilities, we evaluate it on a diverse set of MLLM benchmarks.
As shown in Table~\ref{tab:mllm_result}, \name{} consistently outperforms specialized embodied reasoning models such as Cosmos-Reason1~\citep{azzolini2025cosmos} and RoboBrain2.0~\citep{team2025robobrain} on embodied reasoning tasks, while maintaining strong performance on general MLLM benchmarks. These results indicate that RL-based alignment with low-level actions does not degrade MLLM capabilities; instead, it improves them (See Appendix~\ref{appendix:alignment_setup} for detailed setups).
\vspace{-0.4cm}
\section{Conclusion}
\vspace{-0.2cm}
\label{sec:conclusion} We proposed \name{}, a training framework for MLLMs tailored to vision–language–action models (VLA) by directly aligning MLLM's representations with low-level action policies. 
Our approach leverages reinforcement learning to improve low-level action prediction accuracy through embodied reasoning. 
We evaluated \name{} across diverse robotic environments and MLLM benchmarks, and demonstrated that it consistently delivers substantial gains in embodied reasoning performance within MLLM tasks as well as in the VLA domain across both short and long horizon tasks. 
In contrast, language-only embodied reasoning fine-tuning yields limited or even degraded performance on complex scenarios. 
These results establish \name{} as an effective and generalizable approach for advancing VLA training.

\section*{Impact Statement} This paper proposes a Multimodal Large Language Model (MLLM) training framework that aligns with specific action data. However, if the model is aligned with unsafe action trajectories, Vision-Language-Action Models (VLAs) trained using such MLLMs may be biased toward generating unsafe actions. Since it is difficult to precisely determine on what data an MLLM was trained on after deployment, it is crucial to consider and prevent these issues proactively during the training phase.


\bibliography{example_paper}

@article{zhang2026vlm4vla,
  title={VLM4VLA: Revisiting Vision-Language-Models in Vision-Language-Action Models},
  author={Zhang, Jianke and Chen, Xiaoyu and Wang, Qiuyue and Li, Mingsheng and Guo, Yanjiang and Hu, Yucheng and Zhang, Jiajun and Bai, Shuai and Lin, Junyang and Chen, Jianyu},
  journal={arXiv preprint arXiv:2601.03309},
  year={2026}
}

@article{driess2023palm,
  title={Palm-e: An embodied multimodal language model},
  author={Driess, Danny and Xia, Fei and Sajjadi, Mehdi SM and Lynch, Corey and Chowdhery, Aakanksha and Wahid, Ayzaan and Tompson, Jonathan and Vuong, Quan and Yu, Tianhe and Huang, Wenlong and others},
  journal={arXiv preprint arXiv:2303.03378},
  year={2023}
}

@article{brohan2022rt,
  title={Rt-1: Robotics transformer for real-world control at scale},
  author={Brohan, Anthony and Brown, Noah and Carbajal, Justice and Chebotar, Yevgen and Dabis, Joseph and Finn, Chelsea and Gopalakrishnan, Keerthana and Hausman, Karol and Herzog, Alex and Hsu, Jasmine and others},
  journal={arXiv preprint arXiv:2212.06817},
  year={2022}
}

@article{brohan2023rt,
  title={Rt-2: Vision-language-action models transfer web knowledge to robotic control},
  author={Brohan, Anthony and Brown, Noah and Carbajal, Justice and Chebotar, Yevgen and Chen, Xi and Choromanski, Krzysztof and Ding, Tianli and Driess, Danny and Dubey, Avinava and Finn, Chelsea and others},
  journal={arXiv preprint arXiv:2307.15818},
  year={2023}
}

@article{yang2023learning,
  title={Learning interactive real-world simulators},
  author={Yang, Mengjiao and Du, Yilun and Ghasemipour, Kamyar and Tompson, Jonathan and Schuurmans, Dale and Abbeel, Pieter},
  journal={arXiv preprint arXiv:2310.06114},
  volume={1},
  number={2},
  pages={6},
  year={2023}
}

@article{huang2022inner,
  title={Inner monologue: Embodied reasoning through planning with language models},
  author={Huang, Wenlong and Xia, Fei and Xiao, Ted and Chan, Harris and Liang, Jacky and Florence, Pete and Zeng, Andy and Tompson, Jonathan and Mordatch, Igor and Chebotar, Yevgen and others},
  journal={arXiv preprint arXiv:2207.05608},
  year={2022}
}

@article{tellex2020robots,
  title={Robots that use language},
  author={Tellex, Stefanie and Gopalan, Nakul and Kress-Gazit, Hadas and Matuszek, Cynthia},
  journal={Annual Review of Control, Robotics, and Autonomous Systems},
  volume={3},
  number={1},
  pages={25--55},
  year={2020},
  publisher={Annual Reviews}
}

@article{team2024octo,
  title={Octo: An open-source generalist robot policy},
  author={Team, Octo Model and Ghosh, Dibya and Walke, Homer and Pertsch, Karl and Black, Kevin and Mees, Oier and Dasari, Sudeep and Hejna, Joey and Kreiman, Tobias and Xu, Charles and others},
  journal={arXiv preprint arXiv:2405.12213},
  year={2024}
}

@article{ahn2022can,
  title={Do as i can, not as i say: Grounding language in robotic affordances},
  author={Ahn, Michael and Brohan, Anthony and Brown, Noah and Chebotar, Yevgen and Cortes, Omar and David, Byron and Finn, Chelsea and Fu, Chuyuan and Gopalakrishnan, Keerthana and Hausman, Karol and others},
  journal={arXiv preprint arXiv:2204.01691},
  year={2022}
}

@article{bjorck2025gr00t,
  title={Gr00t n1: An open foundation model for generalist humanoid robots},
  author={Bjorck, Johan and Casta{\~n}eda, Fernando and Cherniadev, Nikita and Da, Xingye and Ding, Runyu and Fan, Linxi and Fang, Yu and Fox, Dieter and Hu, Fengyuan and Huang, Spencer and others},
  journal={arXiv preprint arXiv:2503.14734},
  year={2025}
}

@inproceedings{shentu2024llms,
  title={From LLMs to Actions: latent codes as bridges in hierarchical robot control},
  author={Shentu, Yide and Wu, Philipp and Rajeswaran, Aravind and Abbeel, Pieter},
  booktitle={2024 IEEE/RSJ International Conference on Intelligent Robots and Systems (IROS)},
  pages={8539--8546},
  year={2024},
  organization={IEEE}
}

@article{black2024pi_0,
  title={$\pi_0 $: A Vision-Language-Action Flow Model for General Robot Control},
  author={Black, Kevin and Brown, Noah and Driess, Danny and Esmail, Adnan and Equi, Michael and Finn, Chelsea and Fusai, Niccolo and Groom, Lachy and Hausman, Karol and Ichter, Brian and others},
  journal={arXiv preprint arXiv:2410.24164},
  year={2024}
}

@article{ravi2024sam,
  title={Sam 2: Segment anything in images and videos},
  author={Ravi, Nikhila and Gabeur, Valentin and Hu, Yuan-Ting and Hu, Ronghang and Ryali, Chaitanya and Ma, Tengyu and Khedr, Haitham and R{\"a}dle, Roman and Rolland, Chloe and Gustafson, Laura and others},
  journal={arXiv preprint arXiv:2408.00714},
  year={2024}
}

@article{khazatsky2024droid,
  title={Droid: A large-scale in-the-wild robot manipulation dataset},
  author={Khazatsky, Alexander and Pertsch, Karl and Nair, Suraj and Balakrishna, Ashwin and Dasari, Sudeep and Karamcheti, Siddharth and Nasiriany, Soroush and Srirama, Mohan Kumar and Chen, Lawrence Yunliang and Ellis, Kirsty and others},
  journal={arXiv preprint arXiv:2403.12945},
  year={2024}
}

@inproceedings{walke2023bridgedata,
  title={Bridgedata v2: A dataset for robot learning at scale},
  author={Walke, Homer Rich and Black, Kevin and Zhao, Tony Z and Vuong, Quan and Zheng, Chongyi and Hansen-Estruch, Philippe and He, Andre Wang and Myers, Vivek and Kim, Moo Jin and Du, Max and others},
  booktitle={Conference on Robot Learning},
  pages={1723--1736},
  year={2023},
  organization={PMLR}
}

@article{kim2024openvla,
  title={Openvla: An open-source vision-language-action model},
  author={Kim, Moo Jin and Pertsch, Karl and Karamcheti, Siddharth and Xiao, Ted and Balakrishna, Ashwin and Nair, Suraj and Rafailov, Rafael and Foster, Ethan and Lam, Grace and Sanketi, Pannag and others},
  journal={arXiv preprint arXiv:2406.09246},
  year={2024}
}

@article{pertsch2025fast,
  title={Fast: Efficient action tokenization for vision-language-action models},
  author={Pertsch, Karl and Stachowicz, Kyle and Ichter, Brian and Driess, Danny and Nair, Suraj and Vuong, Quan and Mees, Oier and Finn, Chelsea and Levine, Sergey},
  journal={arXiv preprint arXiv:2501.09747},
  year={2025}
}

@inproceedings{huang2022language,
  title={Language models as zero-shot planners: Extracting actionable knowledge for embodied agents},
  author={Huang, Wenlong and Abbeel, Pieter and Pathak, Deepak and Mordatch, Igor},
  booktitle={International conference on machine learning},
  pages={9118--9147},
  year={2022},
  organization={PMLR}
}

@book{muller2007information,
  title={Information retrieval for music and motion},
  author={M{\"u}ller, Meinard},
  year={2007},
  publisher={Springer}
}

@article{black2025pi0,
  title={$\pi$0. 5: a vision-language-action model with open-world generalization},
  author={Black, Kevin and Brown, Noah and Darpinian, James and Dhabalia, Karan and Driess, Danny and Esmail, Adnan and Equi, Michael and Finn, Chelsea and Fusai, Niccolo and Galliker, Manuel Y and others},
  journal={arXiv preprint arXiv:2504.16054},
  year={2025}
}

@article{driess2025knowledge,
  title={Knowledge insulating vision-language-action models: Train fast, run fast, generalize better},
  author={Driess, Danny and Springenberg, Jost Tobias and Ichter, Brian and Yu, Lili and Li-Bell, Adrian and Pertsch, Karl and Ren, Allen Z and Walke, Homer and Vuong, Quan and Shi, Lucy Xiaoyang and others},
  journal={arXiv preprint arXiv:2505.23705},
  year={2025}
}

@article{kim2025fine,
  title={Fine-tuning vision-language-action models: Optimizing speed and success},
  author={Kim, Moo Jin and Finn, Chelsea and Liang, Percy},
  journal={arXiv preprint arXiv:2502.19645},
  year={2025}
}

@article{zawalski2024robotic,
  title={Robotic control via embodied chain-of-thought reasoning},
  author={Zawalski, Micha{\l} and Chen, William and Pertsch, Karl and Mees, Oier and Finn, Chelsea and Levine, Sergey},
  journal={arXiv preprint arXiv:2407.08693},
  year={2024}
}

@article{chen2025training,
  title={Training Strategies for Efficient Embodied Reasoning},
  author={Chen, William and Belkhale, Suneel and Mirchandani, Suvir and Mees, Oier and Driess, Danny and Pertsch, Karl and Levine, Sergey},
  journal={arXiv preprint arXiv:2505.08243},
  year={2025}
}

@inproceedings{huang2024egoexolearn,
  title={Egoexolearn: A dataset for bridging asynchronous ego-and exo-centric view of procedural activities in real world},
  author={Huang, Yifei and Chen, Guo and Xu, Jilan and Zhang, Mingfang and Yang, Lijin and Pei, Baoqi and Zhang, Hongjie and Dong, Lu and Wang, Yali and Wang, Limin and others},
  booktitle={Proceedings of the IEEE/CVF Conference on Computer Vision and Pattern Recognition},
  pages={22072--22086},
  year={2024}
}

@article{chen2023egoplan,
  title={Egoplan-bench: Benchmarking egocentric embodied planning with multimodal large language models},
  author={Chen, Yi and Ge, Yuying and Ge, Yixiao and Ding, Mingyu and Li, Bohao and Wang, Rui and Xu, Ruifeng and Shan, Ying and Liu, Xihui},
  journal={CoRR},
  year={2023}
}

@article{hu2023look,
  title={Look before you leap: Unveiling the power of gpt-4v in robotic vision-language planning},
  author={Hu, Yingdong and Lin, Fanqi and Zhang, Tong and Yi, Li and Gao, Yang},
  journal={arXiv preprint arXiv:2311.17842},
  year={2023}
}

@article{li2023vision,
  title={Vision-language foundation models as effective robot imitators},
  author={Li, Xinghang and Liu, Minghuan and Zhang, Hanbo and Yu, Cunjun and Xu, Jie and Wu, Hongtao and Cheang, Chilam and Jing, Ya and Zhang, Weinan and Liu, Huaping and others},
  journal={arXiv preprint arXiv:2311.01378},
  year={2023}
}

@article{xu2025multi,
  title={Multi-spatialmllm: Multi-frame spatial understanding with multi-modal large language models},
  author={Xu, Runsen and Wang, Weiyao and Tang, Hao and Chen, Xingyu and Wang, Xiaodong and Chu, Fu-Jen and Lin, Dahua and Feiszli, Matt and Liang, Kevin J},
  journal={arXiv preprint arXiv:2505.17015},
  year={2025}
}

@article{guo2025deepseek,
  title={Deepseek-r1: Incentivizing reasoning capability in llms via reinforcement learning},
  author={Guo, Daya and Yang, Dejian and Zhang, Haowei and Song, Junxiao and Zhang, Ruoyu and Xu, Runxin and Zhu, Qihao and Ma, Shirong and Wang, Peiyi and Bi, Xiao and others},
  journal={arXiv preprint arXiv:2501.12948},
  year={2025}
}

@article{yuan2025embodied,
  title={Embodied-R1: Reinforced Embodied Reasoning for General Robotic Manipulation},
  author={Yuan, Yifu and Cui, Haiqin and Huang, Yaoting and Chen, Yibin and Ni, Fei and Dong, Zibin and Li, Pengyi and Zheng, Yan and Hao, Jianye},
  journal={arXiv preprint arXiv:2508.13998},
  year={2025}
}

@inproceedings{chen2024spatialvlm,
  title={Spatialvlm: Endowing vision-language models with spatial reasoning capabilities},
  author={Chen, Boyuan and Xu, Zhuo and Kirmani, Sean and Ichter, Brain and Sadigh, Dorsa and Guibas, Leonidas and Xia, Fei},
  booktitle={Proceedings of the IEEE/CVF Conference on Computer Vision and Pattern Recognition},
  pages={14455--14465},
  year={2024}
}

@article{luo2025visual,
  title={Visual embodied brain: Let multimodal large language models see, think, and control in spaces},
  author={Luo, Gen and Yang, Ganlin and Gong, Ziyang and Chen, Guanzhou and Duan, Haonan and Cui, Erfei and Tong, Ronglei and Hou, Zhi and Zhang, Tianyi and Chen, Zhe and others},
  journal={arXiv preprint arXiv:2506.00123},
  year={2025}
}

@article{yuan2024robopoint,
  title={Robopoint: A vision-language model for spatial affordance prediction for robotics},
  author={Yuan, Wentao and Duan, Jiafei and Blukis, Valts and Pumacay, Wilbert and Krishna, Ranjay and Murali, Adithyavairavan and Mousavian, Arsalan and Fox, Dieter},
  journal={arXiv preprint arXiv:2406.10721},
  year={2024}
}

@article{chen2024we,
  title={Are we on the right way for evaluating large vision-language models?},
  author={Chen, Lin and Li, Jinsong and Dong, Xiaoyi and Zhang, Pan and Zang, Yuhang and Chen, Zehui and Duan, Haodong and Wang, Jiaqi and Qiao, Yu and Lin, Dahua and others},
  journal={Advances in Neural Information Processing Systems},
  volume={37},
  pages={27056--27087},
  year={2024}
}

@article{kim2025robot,
  title={Robot-R1: Reinforcement Learning for Enhanced Embodied Reasoning in Robotics},
  author={Kim, Dongyoung and Park, Sumin and Jang, Huiwon and Shin, Jinwoo and Kim, Jaehyung and Seo, Younggyo},
  journal={arXiv preprint arXiv:2506.00070},
  year={2025}
}

@article{zheng2024tracevla,
  title={Tracevla: Visual trace prompting enhances spatial-temporal awareness for generalist robotic policies},
  author={Zheng, Ruijie and Liang, Yongyuan and Huang, Shuaiyi and Gao, Jianfeng and Daum{\'e} III, Hal and Kolobov, Andrey and Huang, Furong and Yang, Jianwei},
  journal={arXiv preprint arXiv:2412.10345},
  year={2024}
}

@article{lee2025molmoact,
  title={Molmoact: Action reasoning models that can reason in space},
  author={Lee, Jason and Duan, Jiafei and Fang, Haoquan and Deng, Yuquan and Liu, Shuo and Li, Boyang and Fang, Bohan and Zhang, Jieyu and Wang, Yi Ru and Lee, Sangho and others},
  journal={arXiv preprint arXiv:2508.07917},
  year={2025}
}

@article{azzolini2025cosmos,
  title={Cosmos-reason1: From physical common sense to embodied reasoning},
  author={Azzolini, Alisson and Bai, Junjie and Brandon, Hannah and Cao, Jiaxin and Chattopadhyay, Prithvijit and Chen, Huayu and Chu, Jinju and Cui, Yin and Diamond, Jenna and Ding, Yifan and others},
  journal={arXiv preprint arXiv:2503.15558},
  year={2025}
}

@inproceedings{ji2025robobrain,
  title={Robobrain: A unified brain model for robotic manipulation from abstract to concrete},
  author={Ji, Yuheng and Tan, Huajie and Shi, Jiayu and Hao, Xiaoshuai and Zhang, Yuan and Zhang, Hengyuan and Wang, Pengwei and Zhao, Mengdi and Mu, Yao and An, Pengju and others},
  booktitle={Proceedings of the Computer Vision and Pattern Recognition Conference},
  pages={1724--1734},
  year={2025}
}

@inproceedings{song2025robospatial,
  title={Robospatial: Teaching spatial understanding to 2d and 3d vision-language models for robotics},
  author={Song, Chan Hee and Blukis, Valts and Tremblay, Jonathan and Tyree, Stephen and Su, Yu and Birchfield, Stan},
  booktitle={Proceedings of the Computer Vision and Pattern Recognition Conference},
  pages={15768--15780},
  year={2025}
}

@article{ray2024sat,
  title={Sat: Spatial aptitude training for multimodal language models},
  author={Ray, Arijit and Duan, Jiafei and Tan, Reuben and Bashkirova, Dina and Hendrix, Rose and Ehsani, Kiana and Kembhavi, Aniruddha and Plummer, Bryan A and Krishna, Ranjay and Zeng, Kuo-Hao and others},
  journal={arXiv e-prints},
  pages={arXiv--2412},
  year={2024}
}

@inproceedings{lu2023vl,
  title={Vl-grasp: a 6-dof interactive grasp policy for language-oriented objects in cluttered indoor scenes},
  author={Lu, Yuhao and Fan, Yixuan and Deng, Beixing and Liu, Fangfu and Li, Yali and Wang, Shengjin},
  booktitle={2023 IEEE/RSJ International Conference on Intelligent Robots and Systems (IROS)},
  pages={976--983},
  year={2023},
  organization={IEEE}
}

@article{ranasinghe2024understanding,
  title={Understanding long videos with multimodal language models},
  author={Ranasinghe, Kanchana and Li, Xiang and Kahatapitiya, Kumara and Ryoo, Michael S},
  journal={arXiv preprint arXiv:2403.16998},
  year={2024}
}

@inproceedings{yang2025magma,
  title={Magma: A foundation model for multimodal ai agents},
  author={Yang, Jianwei and Tan, Reuben and Wu, Qianhui and Zheng, Ruijie and Peng, Baolin and Liang, Yongyuan and Gu, Yu and Cai, Mu and Ye, Seonghyeon and Jang, Joel and others},
  booktitle={Proceedings of the Computer Vision and Pattern Recognition Conference},
  pages={14203--14214},
  year={2025}
}

@article{wu2025spatial,
  title={Spatial-mllm: Boosting mllm capabilities in visual-based spatial intelligence},
  author={Wu, Diankun and Liu, Fangfu and Hung, Yi-Hsin and Duan, Yueqi},
  journal={arXiv preprint arXiv:2505.23747},
  year={2025}
}

@article{tong2024cambrian,
  title={Cambrian-1: A fully open, vision-centric exploration of multimodal llms},
  author={Tong, Peter and Brown, Ellis and Wu, Penghao and Woo, Sanghyun and IYER, Adithya Jairam Vedagiri and Akula, Sai Charitha and Yang, Shusheng and Yang, Jihan and Middepogu, Manoj and Wang, Ziteng and others},
  journal={Advances in Neural Information Processing Systems},
  volume={37},
  pages={87310--87356},
  year={2024}
}

@article{zhou2025roborefer,
  title={RoboRefer: Towards Spatial Referring with Reasoning in Vision-Language Models for Robotics},
  author={Zhou, Enshen and An, Jingkun and Chi, Cheng and Han, Yi and Rong, Shanyu and Zhang, Chi and Wang, Pengwei and Wang, Zhongyuan and Huang, Tiejun and Sheng, Lu and others},
  journal={arXiv preprint arXiv:2506.04308},
  year={2025}
}

@article{yuan2025seeing,
  title={From Seeing to Doing: Bridging Reasoning and Decision for Robotic Manipulation},
  author={Yuan, Yifu and Cui, Haiqin and Chen, Yibin and Dong, Zibin and Ni, Fei and Kou, Longxin and Liu, Jinyi and Li, Pengyi and Zheng, Yan and Hao, Jianye},
  journal={arXiv preprint arXiv:2505.08548},
  year={2025}
}

@article{huang2025thinkact,
  title={Thinkact: Vision-language-action reasoning via reinforced visual latent planning},
  author={Huang, Chi-Pin and Wu, Yueh-Hua and Chen, Min-Hung and Wang, Yu-Chiang Frank and Yang, Fu-En},
  journal={arXiv preprint arXiv:2507.16815},
  year={2025}
}

@article{song2025maniplvm,
  title={Maniplvm-r1: Reinforcement learning for reasoning in embodied manipulation with large vision-language models},
  author={Song, Zirui and Ouyang, Guangxian and Li, Mingzhe and Ji, Yuheng and Wang, Chenxi and Xu, Zixiang and Zhang, Zeyu and Zhang, Xiaoqing and Jiang, Qian and Chen, Zhenhao and others},
  journal={arXiv preprint arXiv:2505.16517},
  year={2025}
}

@article{wei2022chain,
  title={Chain-of-thought prompting elicits reasoning in large language models},
  author={Wei, Jason and Wang, Xuezhi and Schuurmans, Dale and Bosma, Maarten and Xia, Fei and Chi, Ed and Le, Quoc V and Zhou, Denny and others},
  journal={Advances in neural information processing systems},
  volume={35},
  pages={24824--24837},
  year={2022}
}

@article{wang2022self,
  title={Self-consistency improves chain of thought reasoning in language models},
  author={Wang, Xuezhi and Wei, Jason and Schuurmans, Dale and Le, Quoc and Chi, Ed and Narang, Sharan and Chowdhery, Aakanksha and Zhou, Denny},
  journal={arXiv preprint arXiv:2203.11171},
  year={2022}
}

@article{yao2023tree,
  title={Tree of thoughts: Deliberate problem solving with large language models},
  author={Yao, Shunyu and Yu, Dian and Zhao, Jeffrey and Shafran, Izhak and Griffiths, Tom and Cao, Yuan and Narasimhan, Karthik},
  journal={Advances in neural information processing systems},
  volume={36},
  pages={11809--11822},
  year={2023}
}

@article{kim2023cot,
  title={The cot collection: Improving zero-shot and few-shot learning of language models via chain-of-thought fine-tuning},
  author={Kim, Seungone and Joo, Se June and Kim, Doyoung and Jang, Joel and Ye, Seonghyeon and Shin, Jamin and Seo, Minjoon},
  journal={arXiv preprint arXiv:2305.14045},
  year={2023}
}

@article{muennighoff2025s1,
  title={s1: Simple test-time scaling},
  author={Muennighoff, Niklas and Yang, Zitong and Shi, Weijia and Li, Xiang Lisa and Fei-Fei, Li and Hajishirzi, Hannaneh and Zettlemoyer, Luke and Liang, Percy and Cand{\`e}s, Emmanuel and Hashimoto, Tatsunori},
  journal={arXiv preprint arXiv:2501.19393},
  year={2025}
}

@article{wang2025reinforcement,
  title={Reinforcement Learning for Reasoning in Large Language Models with One Training Example},
  author={Wang, Yiping and Yang, Qing and Zeng, Zhiyuan and Ren, Liliang and Liu, Lucas and Peng, Baolin and Cheng, Hao and He, Xuehai and Wang, Kuan and Gao, Jianfeng and others},
  journal={arXiv preprint arXiv:2504.20571},
  year={2025}
}

@article{yu2025dapo,
  title={Dapo: An open-source llm reinforcement learning system at scale},
  author={Yu, Qiying and Zhang, Zheng and Zhu, Ruofei and Yuan, Yufeng and Zuo, Xiaochen and Yue, Yu and Fan, Tiantian and Liu, Gaohong and Liu, Lingjun and Liu, Xin and others},
  journal={arXiv preprint arXiv:2503.14476},
  year={2025}
}

@article{ahmed2006discrete,
  title={Discrete cosine transform},
  author={Ahmed, Nasir and Natarajan, T\_ and Rao, Kamisetty R},
  journal={IEEE transactions on Computers},
  volume={100},
  number={1},
  pages={90--93},
  year={2006},
  publisher={IEEE}
}

@article{gage1994new,
  title={A new algorithm for data compression},
  author={Gage, Philip},
  journal={C Users Journal},
  volume={12},
  number={2},
  pages={23--38},
  year={1994},
  publisher={McPherson, KS: R \& D Publications, c1987-1994.}
}

@article{jin2025search,
  title={Search-r1: Training llms to reason and leverage search engines with reinforcement learning},
  author={Jin, Bowen and Zeng, Hansi and Yue, Zhenrui and Yoon, Jinsung and Arik, Sercan and Wang, Dong and Zamani, Hamed and Han, Jiawei},
  journal={arXiv preprint arXiv:2503.09516},
  year={2025}
}

@article{lu2025ui,
  title={Ui-r1: Enhancing action prediction of gui agents by reinforcement learning},
  author={Lu, Zhengxi and Chai, Yuxiang and Guo, Yaxuan and Yin, Xi and Liu, Liang and Wang, Hao and Xiong, Guanjing and Li, Hongsheng},
  journal={arXiv preprint arXiv:2503.21620},
  year={2025}
}

@article{huang2025vision,
  title={Vision-r1: Incentivizing reasoning capability in multimodal large language models},
  author={Huang, Wenxuan and Jia, Bohan and Zhai, Zijie and Cao, Shaosheng and Ye, Zheyu and Zhao, Fei and Xu, Zhe and Hu, Yao and Lin, Shaohui},
  journal={arXiv preprint arXiv:2503.06749},
  year={2025}
}

@article{zeng2025simplerl,
  title={Simplerl-zoo: Investigating and taming zero reinforcement learning for open base models in the wild},
  author={Zeng, Weihao and Huang, Yuzhen and Liu, Qian and Liu, Wei and He, Keqing and Ma, Zejun and He, Junxian},
  journal={arXiv preprint arXiv:2503.18892},
  year={2025}
}

@article{shen2025vlm,
  title={Vlm-r1: A stable and generalizable r1-style large vision-language model},
  author={Shen, Haozhan and Liu, Peng and Li, Jingcheng and Fang, Chunxin and Ma, Yibo and Liao, Jiajia and Shen, Qiaoli and Zhang, Zilun and Zhao, Kangjia and Zhang, Qianqian and others},
  journal={arXiv preprint arXiv:2504.07615},
  year={2025}
}

@article{shao2024deepseekmath,
  title={Deepseekmath: Pushing the limits of mathematical reasoning in open language models},
  author={Shao, Zhihong and Wang, Peiyi and Zhu, Qihao and Xu, Runxin and Song, Junxiao and Bi, Xiao and Zhang, Haowei and Zhang, Mingchuan and Li, YK and Wu, Y and others},
  journal={arXiv preprint arXiv:2402.03300},
  year={2024}
}

@article{mees2022calvin,
  title={Calvin: A benchmark for language-conditioned policy learning for long-horizon robot manipulation tasks},
  author={Mees, Oier and Hermann, Lukas and Rosete-Beas, Erick and Burgard, Wolfram},
  journal={IEEE Robotics and Automation Letters},
  volume={7},
  number={3},
  pages={7327--7334},
  year={2022},
  publisher={IEEE}
}

@article{liu2023libero,
  title={Libero: Benchmarking knowledge transfer for lifelong robot learning},
  author={Liu, Bo and Zhu, Yifeng and Gao, Chongkai and Feng, Yihao and Liu, Qiang and Zhu, Yuke and Stone, Peter},
  journal={Advances in Neural Information Processing Systems},
  volume={36},
  pages={44776--44791},
  year={2023}
}

@misc{nvidia2025gr00t,
  author       = {NVIDIA GEAR},
  title        = {GR00T N1.5: An Improved Open Foundation Model for Generalist Humanoid Robots},
  howpublished = {\url{https://research.nvidia.com/labs/gear/gr00t-n1_5/}},
  note         = {Accessed: 2025-09-09},
  year         = {2025},
  month        = {June},
  day          = {11}
}

@inproceedings{sermanet2024robovqa,
  title={Robovqa: Multimodal long-horizon reasoning for robotics},
  author={Sermanet, Pierre and Ding, Tianli and Zhao, Jeffrey and Xia, Fei and Dwibedi, Debidatta and Gopalakrishnan, Keerthana and Chan, Christine and Dulac-Arnold, Gabriel and Maddineni, Sharath and Joshi, Nikhil J and others},
  booktitle={2024 IEEE International Conference on Robotics and Automation (ICRA)},
  pages={645--652},
  year={2024},
  organization={IEEE}
}

@inproceedings{liang2023code,
  title={Code as policies: Language model programs for embodied control},
  author={Liang, Jacky and Huang, Wenlong and Xia, Fei and Xu, Peng and Hausman, Karol and Ichter, Brian and Florence, Pete and Zeng, Andy},
  booktitle={2023 IEEE International Conference on Robotics and Automation (ICRA)},
  pages={9493--9500},
  year={2023},
  organization={IEEE}
}

@inproceedings{fu2024blink,
  title={Blink: Multimodal large language models can see but not perceive},
  author={Fu, Xingyu and Hu, Yushi and Li, Bangzheng and Feng, Yu and Wang, Haoyu and Lin, Xudong and Roth, Dan and Smith, Noah A and Ma, Wei-Chiu and Krishna, Ranjay},
  booktitle={European Conference on Computer Vision},
  pages={148--166},
  year={2024},
  organization={Springer}
}

@article{li2024llava,
  title={Llava-onevision: Easy visual task transfer},
  author={Li, Bo and Zhang, Yuanhan and Guo, Dong and Zhang, Renrui and Li, Feng and Zhang, Hao and Zhang, Kaichen and Zhang, Peiyuan and Li, Yanwei and Liu, Ziwei and others},
  journal={arXiv preprint arXiv:2408.03326},
  year={2024}
}

@article{cheng2024spatialrgpt,
  title={SpatialRGPT: Grounded Spatial Reasoning in Vision Language Models},
  author={Cheng, An-Chieh and Yin, Hongxu and Fu, Yang and Guo, Qiushan and Yang, Ruihan and Kautz, Jan and Wang, Xiaolong and Liu, Sifei},
  journal={arXiv preprint arXiv:2406.01584},
  year={2024}
}

@article{bai2025qwen2,
  title={Qwen2. 5-vl technical report},
  author={Bai, Shuai and Chen, Keqin and Liu, Xuejing and Wang, Jialin and Ge, Wenbin and Song, Sibo and Dang, Kai and Wang, Peng and Wang, Shijie and Tang, Jun and others},
  journal={arXiv preprint arXiv:2502.13923},
  year={2025}
}

@article{lynch2023interactive,
  title={Interactive language: Talking to robots in real time},
  author={Lynch, Corey and Wahid, Ayzaan and Tompson, Jonathan and Ding, Tianli and Betker, James and Baruch, Robert and Armstrong, Travis and Florence, Pete},
  journal={IEEE Robotics and Automation Letters},
  year={2023},
  publisher={IEEE}
}

@inproceedings{zhao2025cot,
  title={Cot-vla: Visual chain-of-thought reasoning for vision-language-action models},
  author={Zhao, Qingqing and Lu, Yao and Kim, Moo Jin and Fu, Zipeng and Zhang, Zhuoyang and Wu, Yecheng and Li, Zhaoshuo and Ma, Qianli and Han, Song and Finn, Chelsea and others},
  booktitle={Proceedings of the Computer Vision and Pattern Recognition Conference},
  pages={1702--1713},
  year={2025}
}

@article{team2025robobrain,
  title={Robobrain 2.0 technical report},
  author={Team, BAAI RoboBrain and Cao, Mingyu and Tan, Huajie and Ji, Yuheng and Lin, Minglan and Li, Zhiyu and Cao, Zhou and Wang, Pengwei and Zhou, Enshen and Han, Yi and others},
  journal={arXiv preprint arXiv:2507.02029},
  year={2025}
}

@article{openai2024gpt4o,
  title={hello-gpt-4o},
  author={OpenAI},
  journal={OpenAI Blog},
  year={2024},
  month={May},
  url={https://openai.com/index/hello-gpt-4o/}
}

@article{qwen2025qwen3vl,
  title={Qwen3-VL: Sharper Vision, Deeper Thought, Broader Action},
  author={Qwen Team},
  journal={Qwen Blog},
  year={2025},
  month={September},
  url={https://qwen.ai/blog?id=99f0335c4ad9ff6153e517418d48535ab6d8afef&from=research.latest-advancements-list}
}

@article{googledeepMind2025geminiUpdate,
  title={Gemini Model Thinking Updates March 2025},
  author={Hassabis, Demis and Kavukcuoglu, Koray and Google DeepMind},
  journal={Google Blog},
  year={2025},
  month={March},
  url={https://blog.google/technology/google-deepmind/gemini-model-thinking-updates-march-2025/}
}

@article{hurst2024gpt,
  title={Gpt-4o system card},
  author={Hurst, Aaron and Lerer, Adam and Goucher, Adam P and Perelman, Adam and Ramesh, Aditya and Clark, Aidan and Ostrow, AJ and Welihinda, Akila and Hayes, Alan and Radford, Alec and others},
  journal={arXiv preprint arXiv:2410.21276},
  year={2024}
}

@inproceedings{duan2024vlmevalkit,
  title={Vlmevalkit: An open-source toolkit for evaluating large multi-modality models},
  author={Duan, Haodong and Yang, Junming and Qiao, Yuxuan and Fang, Xinyu and Chen, Lin and Liu, Yuan and Dong, Xiaoyi and Zang, Yuhang and Zhang, Pan and Wang, Jiaqi and others},
  booktitle={Proceedings of the 32nd ACM International Conference on Multimedia},
  pages={11198--11201},
  year={2024}
}
\bibliographystyle{icml2026}

\newpage
\appendix
\onecolumn

\section{Experiment Details}
\label{appendix:detail} 

\subsection{Computing Cost} 
We use 8×H200 GPUs for MLLM training, requiring approximately 30 hours for SFT and 1 hour for reinforcement learning. For VLA training, we use 2×A100 GPUs, with each 10K training steps taking about 1 hour of computation.  

\subsection{Implementation Details for VLA Training} 
Our implementation refers to the GR00T-N1.5 codebase\footnote{https://github.com/NVIDIA/Isaac-GR00T}~\citep{nvidia2025gr00t}, adopting the same architecture with an initialized diffusion policy action expert. 
The expert takes as input the hidden states from the 18-th layer of Qwen2.5VL-7B-Ins~\citep{bai2025qwen2}. Hyperparameters for policy fine-tuning follow those of the official GR00T-N1.5 implementation unless otherwise specified. Action experts are newly trained for each benchmark environment with a batch size of 32. 
Training steps are set to 60K for LIBERO, 100K for CALVIN, 30K for real robot.

\subsection{Training Datasets} For supervised fine-tuning (SFT), we prepare a diverse set of datasets covering both general MLLM capability and embodied reasoning. To preserve general multimodal ability, we use 100K samples from LLaVA-OneVision (single-view only)~\citep{li2024llava}. For embodied reasoning, we include 300K samples from RefSpatial~\citep{zhou2025roborefer}, 200K from RoboPoint~\citep{yuan2024robopoint}, 50K from EgoPlan-IT~\citep{chen2023egoplan}, and 500K from our own multi-view instruction dataset. To enhance robot-specific embodied reasoning, we incorporate 100K samples each from ShareRobot~\citep{ji2025robobrain} and RobotVQA~\citep{sermanet2024robovqa}, 150K from our \name{} VQA, and 300K from BridgeV2~\citep{walke2023bridgedata} and Droid~\citep{khazatsky2024droid} Robot QA (predicting movements such as “move right,” “move left,” the current 7-DoF state, and a future sequence of 10 states). Since conventional robot imitation environments do not take video inputs, video-based datasets (RobotVQA, EgoPlan-IT, ShareRobot) are converted into single-frame inputs by extracting the last frame. For reasoning data, we include 50K multiple-choice QA samples converted from our \name{} VQA dataset and another 50K derived from SAM2~\citep{ravi2024sam}, which queries spatial relations among key objects. Of these, 30K samples are used to train the reasoning distillation model. After augmenting with generated data and applying correctness filtering, the final reasoning dataset consists of 76K samples.  In total, the MLLM training set contains 1.88M QA samples. For FAST token prediction~\citep{pertsch2025fast}, we use the subset of BridgeV2 dataset (400K samples). For reinforcement learning, we further use a 12.8K subset of the BridgeV2 FAST token prediction data. The training for FAST token prediction follows the prompt template shown in Figure~\ref{fig:fast_prompt}.

\subsection{Analysis Setup Details}\label{appendix:alignment_setup}

\textbf{Comparison with embodied alignment strategies.} For a fair comparison, all models are trained with RL on the same BridgeV2 images as \name{}, encouraging embodied reasoning in both cases. 
For high-level action alignment, we convert movements such as "\textit{move right}" or "\textit{move left}" into multiple-choice QA format and provide rewards based on  correctness~\citep{kim2025robot}. For 2D visual trajectory prediction, we use data from ShareRobot and adopt the same reward formulation as ThinkAct~\citep{huang2025thinkact}. Since ShareRobot contains only 6K samples, training is limited to this size. After training, all models are converted into VLAs and evaluated on the LIBERO simulation environments.

\textbf{Comparison with SFT-based alignment. } For this experiment, we use the ECoT dataset while keeping the action space in the form of FAST tokens. Both methods are trained on the same 12.8K samples on top of \name{} SFT model, with one epoch of SFT using identical hyperparameters. Then, the resulting models are converted into VLAs and evaluated on the LIBERO simulation environments.

\textbf{Representation analysis.  }  For this experiment, We select 20 training trajectories from single LIBERO long-horizon task, "put the white mug on the left plate and put the yellow and white mug on the right plate." 
We assign each timestep to 32 classes using Dynamic Time Warping (DTW)~\citep{muller2007information} over robot states. 
We then evaluate whether the MLLM, receiving only vision and task instruction, can recover the correct underlying class using a KNN classifier ($k=5$) applied to its hidden representation.

\newcommand{\refvala}[1]{\phantom{$^{*}$}#1$^{*}$}
\newcommand{\refvalb}[1]{\phantom{$^{\dagger}$}#1$^{\dagger}$}

\begin{table*}[t]
\centering
\caption{\textbf{Performance on multimodal benchmarks.} We evaluate \name{} and other MLLMs on general image understanding (MMStar), spatial reasoning (RoboSpatial, Where2Place, BLINK, and robot embodied reasoning (Robot-R1 Bench) benchmarks. 
Our initial SFT model, \name{} (SFT), performs on par with specialized embodied-reasoning MLLMs, and RL training further boosts performance across the overall MLLM benchmarks. Values marked with $*$ are taken from prior work \citep{team2025robobrain, duan2024vlmevalkit}.
}
\vspace{-0.1in}
\label{tab:mllm_result}
\begin{adjustbox}{max width=0.95\textwidth}
\begin{tabular}{l|ccccc}
\toprule
Model & MMStar & Robot-R1 Bench (0-3) & RoboSpatial & Where2Place & Blink (Rel. Depth) \\
\midrule
GPT-4o-2024-11-20~\citep{hurst2024gpt}                 & \refvala{65.10} & 1.55             & \refvala{44.42} & \refvala{20.41} & \refvala{77.90} \\ 
\midrule
Qwen2.5-VL-7B-Ins~\citep{bai2025qwen2} & 60.30 & 1.02             & 36.29          & 11.35          & 55.64 \\
Cosmos-Reason1-7B~\citep{azzolini2025cosmos}      & 54.40 & \underline{1.19} & \refvala{38.81} & \refvala{\phantom{0}5.51} & \refvala{68.57} \\
RoboBrain2.0-7B~\citep{team2025robobrain}        & 35.80 & 1.17             & \refvala{\textbf{54.23}} & \refvala{\textbf{63.59}} & \refvala{83.95} \\
VeBrain-8B~\citep{luo2025visual}             & 61.90 & 1.02             & \refvala{42.48} & \refvala{11.34} & \refvala{79.68} \\

\rowcolor{gray!20} \name{} w/o RL & 62.47 & 1.14 & 48.86 & 51.66 & \underline{87.10} \\
\rowcolor{gray!20} \name{} (Ours) & \textbf{62.80} & \textbf{1.38} & \underline{50.86} & \underline{54.49} & \textbf{87.90} \\
\bottomrule
\end{tabular}
\end{adjustbox}
\end{table*}

\textbf{Performance on MLLM Benchmarks.}
We evaluate \name{} on a diverse set of MLLM benchmarks to assess both generalist capabilities and embodied reasoning performance.
For general visual question answering (VQA), we use MMStar.
For spatial reasoning, we adopt Robospatial-Home, Where2Place, and the depth-related components of BLINK, which evaluate spatial understanding and geometric reasoning.
For robot embodied reasoning, we use Robot-R1 Bench, which provides fine-grained evaluations of embodied reasoning abilities including planning, subtask decomposition, movement understanding, and spatial reasoning, all constructed on the BridgeV2 dataset. All benchmarks are evaluated following their official protocols, and detailed evaluation settings are reported in Table~\ref{tab:mllm_result}.

\begin{figure*}[ht]
    \centering
    \begin{tcolorbox}[
      title=Waypoint prediction QA for SFT,
      colback=gray!5!white,
      colframe=gray!40!black,
      fonttitle=\bfseries,
      rounded corners,
      width=0.95\linewidth,
      boxrule=0.4pt,
    ]
    \footnotesize
    \begin{center}
    \textbf{System Prompt}
    \end{center}
    You are an embodied vision-language robotic assistant for multi-object manipulation. 
    \par\noindent\rule{\linewidth}{0.4pt}
    \begin{center}
    \textbf{System Prompt for Reinforcement Learning}
    \end{center}
    You are an embodied vision-language robotic assistant for multi-object manipulation. The assistant first thinks about the reasoning process in the mind and then provides the user with the answer. The reasoning process and answer are enclosed within <think> </think> and <answer> </answer> tags, respectively. 
    \par\noindent\rule{\linewidth}{0.4pt}
    \begin{center}
    \textbf{Prompt}
    \end{center}
    Your current task is {instruction}. Output the robot's actions to perform this task through FAST tokens.
    \end{tcolorbox}
    \caption{\textbf{Prompt for FAST token generation.} We use this prompt template for both FAST token prediction and reinforcement learning.}
    \label{fig:fast_prompt}
\end{figure*}

\label{appendix:qaul}

\subsection{RL Training Process} 
\begin{figure}[htbp]
    \centering
    \begin{subfigure}[b]{0.48\textwidth}
        \centering
        \includegraphics[width=\textwidth]{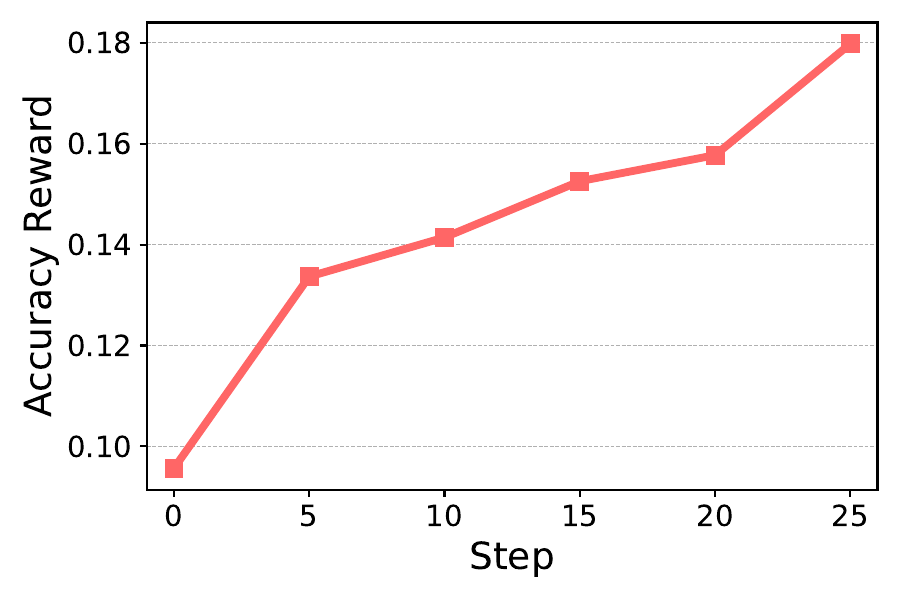}
        \caption{Accuracy reward change during RL training}
        \label{fig:accuracy}
    \end{subfigure}
    \hfill
    \begin{subfigure}[b]{0.48\textwidth}
        \centering
        \includegraphics[width=\textwidth]{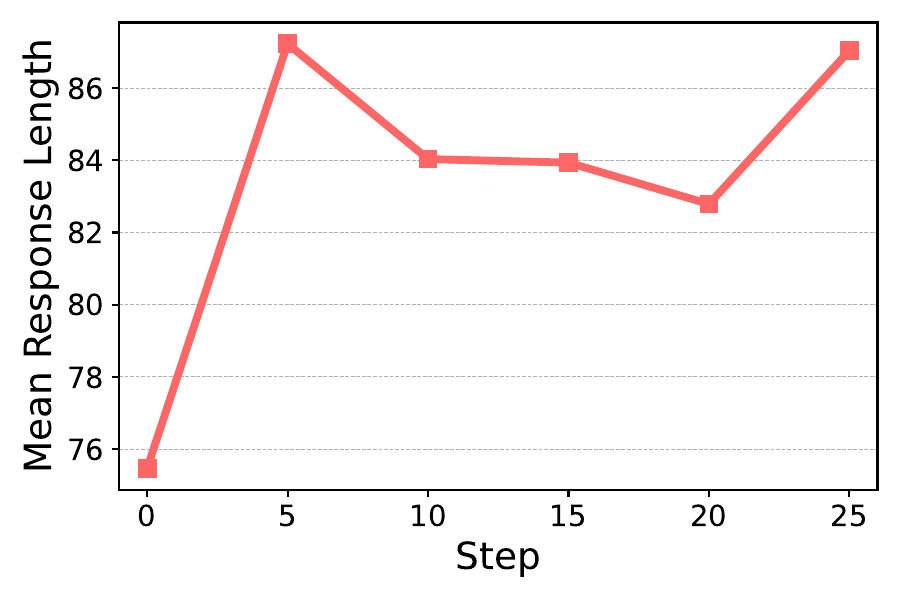}
        \caption{Response length change during RL training}
        \label{fig:length}
    \end{subfigure}
    \caption{\textbf{Training metrics for reinforcement learning in \name{}.} 
    Validation performance during RL training: (a) accuracy reward increases steadily, while (b) response length remains stable.}

    \label{fig:comparison}
\end{figure}

Figures~\ref{fig:comparison} and 2 illustrate how the average response length and accuracy reward evolve during the RL training of \name{}. 
The average response length remains stable, while the accuracy reward gradually increases, indicating effective learning.  

Figure~\ref{fig:rl_reponse_exmplae1} provides qualitative examples before and after RL training. Prior to training, responses mainly describe high-level plans with minimal detail on immediate actions (e.g., “the robot needs to first approach the cloth”). After RL training, the model generates more concrete action descriptions, such as “the robot needs to first approach and align its gripper with the cloth. Once aligned, it will need to close the gripper to pick up the cloth.”. In addition, a novel element of situational awareness has been introduced (e.g., "The robot’s current position suggests it is already aligned with the cloth, ready to pick it up.").
These quantitative and qualitative results together demonstrate that RL training is successfully improving the model’s ability to align reasoning with actionable steps.

\begin{figure}[ht]
    \centering
    \small
    \begin{tabular}{p{0.95\linewidth}}
    \toprule 
        \textbf{Question:} Your current task is move the orange cloth to the top of the right burner. Output the robot's actions to perform this task through FAST tokens.  \\

\midrule 
\textbf{GT Answer:} \\
<|action\_start|><|action\_486|><|action\_265|><|action\_268|><|action\_116|><|action\_269|><|action\_end|> \\
\midrule 
\textbf{\name{} (SFT) Answer:} \\
'<think>To perform the task of moving the orange cloth to the top of the right burner, the robot needs to first approach the cloth, then grasp it, and finally move it to the right burner. The actions should be sequential and purposeful, focusing on the cloth and the burner in question.</think><answer>\\
<|action\_start|><|action\_266|><|action\_709|><|action\_268|><|action\_116|><|action\_269|><|action\_end|>
\\ \midrule 
Reward: 0.5 (Format 1.0 / Accuracy 0.0) \\
\midrule 
\textbf{\name{} (SFT+RL) Answer:} <think>To perform the task of moving the orange cloth to the top of the right burner, the robot needs to first approach and align its gripper with the cloth. Once aligned, it will need to close the gripper to pick up the cloth, lift it, and then move it to the right burner before releasing it. {\color{blue} The robot's current position suggests it is already aligned with the cloth, ready to pick it up.}</think><answer><|action\_start|><|action\_486|><|action\_265|><|action\_268|><|action\_116|><|action\_269|><|action\_end|>\\

\midrule 
Reward: 1.0  (Format 1.0 / Accuracy 1.0)\\
\bottomrule
    \end{tabular}
    \caption{\textbf{Example of model response.}}
    \label{fig:rl_reponse_exmplae1}
\end{figure}

\end{document}